\documentclass[10pt,twocolumn,letterpaper]{article}

\usepackage{cvpr}
\usepackage{times}
\usepackage{epsfig}
\usepackage{graphicx}
\usepackage{amsmath}
\usepackage{amssymb}
\usepackage{float}
\usepackage{multirow}
\usepackage{caption}
\usepackage{subfiles}
\usepackage{lipsum}
\usepackage{pdfpages}
\newcommand\blfootnote[1]{%
  \begingroup
  \renewcommand\thefootnote{}\footnote{#1}%
  \addtocounter{footnote}{-1}%
  \endgroup
}

\usepackage[pagebackref=true,breaklinks=true,letterpaper=true,colorlinks,bookmarks=false]{hyperref}

\cvprfinalcopy 

\def\cvprPaperID{7884} 

\ifcvprfinal\fi
\begin{document}

\title{Compressed Volumetric Heatmaps for Multi-Person 3D Pose Estimation}

\author{
    \textsuperscript{1}University of Modena and Reggio Emilia\\[-1px]
    {\tt\small \{name.surname\}@unimore.it}
\and
    \textsuperscript{2}Panasonic R\&D Company of America\\[-1px]
    {\tt\small \{name.surname\}@us.panasonic.com}
}

\maketitle

\blfootnote{\scriptsize * Work done while interning at Panasonic R\&D Company of America}

\begin{abstract}
In this paper we present a novel approach for bottom-up multi-person 3D human pose estimation from monocular RGB images. We propose to use high resolution volumetric heatmaps to model joint locations, devising a simple and effective compression method to drastically reduce the size of this representation.
At the core of the proposed method lies our Volumetric Heatmap Autoencoder, a fully-convolutional network tasked with the compression of ground-truth heatmaps into a dense intermediate representation. A second model, the Code Predictor, is then trained to predict these codes, which can be decompressed at test time to re-obtain the original representation.
Our experimental evaluation shows that our method performs favorably when compared to state of the art on both multi-person and single-person 3D human pose estimation datasets and, thanks to our novel compression strategy, can process full-HD images at the constant runtime of 8 fps regardless of the number of subjects in the scene. 
Code and models available at \url{https://github.com/fabbrimatteo/LoCO}.




\end{abstract}

\section{Introduction}
\noindent
Human Pose Estimation (HPE) has seen significant progress in recent years, mainly thanks to deep Convolutional Neural Networks (CNNs). Best performing methods on 2D HPE are all leveraging heatmaps to predict body joint locations \cite{cao2017realtime, simple_baselines, tang2018deeply}. Heatmaps have also been extended for 3D HPE, showing promising results in single person contexts \cite{sarandi2018synthetic, pavlakos2017coarse, sun2017integral}.

Despite their good performance, these methods do not easily generalize to multi-person 3D HPE, mainly because of their high demands for memory and computation. This drawback also limits the resolution of those maps, that have to be kept small, leading to quantization errors. Using larger volumetric heatmaps can address those issues, but at the cost of extra storage, computation and training complexity.

\begin{figure}[t!]
\begin{center}
\includegraphics[width=0.95\linewidth]{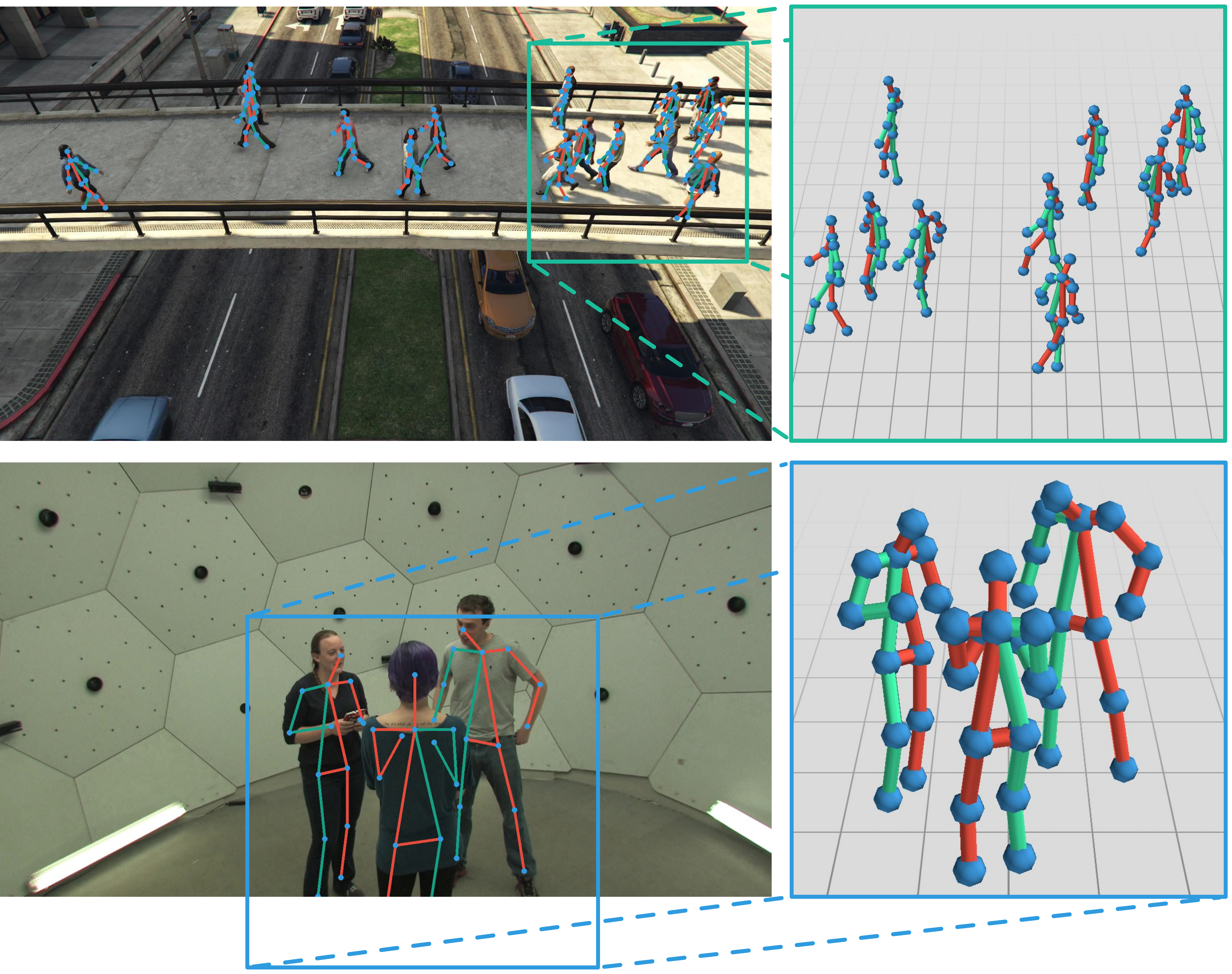}
\end{center}
    \vspace{-10px}
    \caption{Examples of 3D poses estimated by our LoCO approach. Close-ups show that 3D poses are correctly computed even in very complex and articulated scenarios}
\label{fig:cover}
\end{figure}

In this paper, we propose a simple solution to the aforementioned problems that allows us to directly predict high-resolution volumetric heatmaps while keeping storage and computation small. This new solution enables our method to tackle multi-person 3D HPE using heatmaps in a single-shot bottom-up fashion. Moreover, thanks to our high-resolution output, we are able to produce fine-grained absolute 3D predictions even in single person contexts. This allows our method to achieve state of the art performance on the most popular single person benchmark \cite{h36m_pami}.

The core of our proposal relies on the creation of an alternative ground-truth representation that preserves the most informative content of the original ground-truth but reduces its memory footprint. Indeed, this new compressed representation is used as the target ground-truth during our network training. We named this solution LoCO, \emph{Learning on Compressed Output}.

By leveraging on the analogy between compression and dimensionality reduction on sparse signals \cite{WANG2012120,10.1007/978-3-540-73750-6_2,8513469}, we empirically follow the intuition that 3D body poses can be represented in an alternative space where data redundancy is exploited towards a compact representation. This is done by minimizing the loss of information while keeping the spatial nature of the representation, a task for which convolutional architectures are particularly suitable. Concurrently w.r.t. our proposal, compression-based approaches have been effectively used for both dataset distillation and input compression \cite{wang2019dataset,torfason2018towards} but, to the best of our knowledge, this is the first time they are applied to ground truth remapping. For this purpose, deep self-supervised networks such as autoencoders represent a natural choice for searching, in a data-driven way, for an intermediate representation. 

Specifically, our HPE pipeline consists of two modules: at first, the pretrained \textit{Volumetric Heatmap Autoencoder} is used to obtain a smaller/denser representation of the volumetric heatmaps. These ``codes" are then used to supervise the \textit{Code Predictor}, which aims at estimating multiple 3D joint locations from a monocular RGB input.

To summarize, the novel aspects of our proposal are:
\begin{itemize} 
\item We propose a simple and effective method that maps high-resolution volumetric heatmaps to a compact and more tractable representation. This saves memory and computational resources while keeping most of the informative content. 
\item This new data representation enables the adoption of volumetric heatmaps to tackle multi-person 3D HPE in a bottom-up fashion, an otherwise intractable problem. 
Experiments on both real \cite{panoptic_1} and simulated environments \cite{fabbri2018learning} (see Fig.~\ref{fig:cover}) show promising results even in 100 meters wide scenes with more than 50 people. Our method only requires a single forward pass and can be applied with constant running time regardless of the number of subjects in the scene.
\item We further demonstrate the generalization capabilities of LoCO by applying it to a single person context. Our fine-grained predictions establish a new state of the art on Human3.6m \cite{h36m_pami} among bottom-up methods.
\end{itemize}


\begin{figure*}[t!]
\begin{center}
\vspace{-0.5em}
\includegraphics[width=\linewidth]{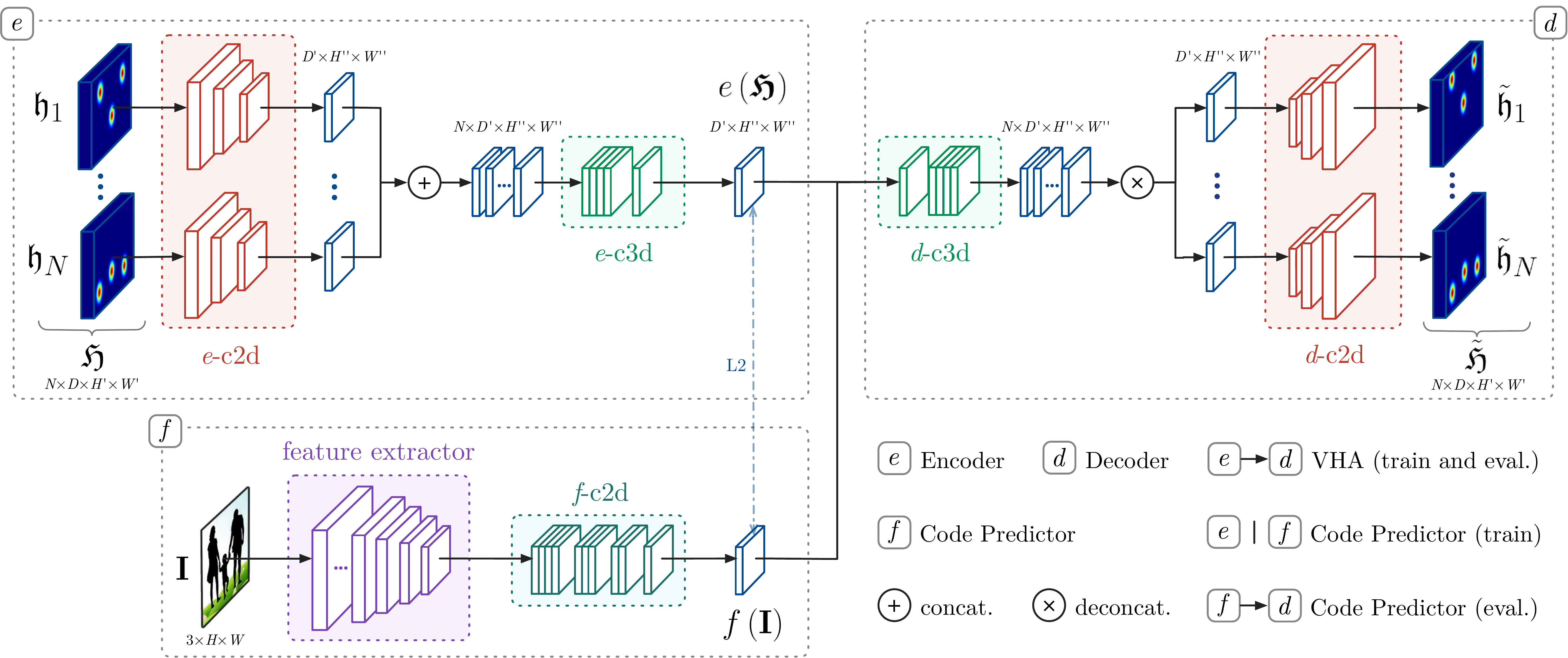}
\end{center}
    \vspace{-0.5em}
    \caption{Schematization of the proposed LoCO pipeline. At training time, the Encoder $e$ produces the compressed volumetric heatmaps $e(\mathfrak{H})$ which are used as ground truth from the Code Predictor $f$. At test time, the intermediate representation $f(I)$ computed by the Code Predictor is fed to the Decoder $d$ for the final output. In our case, $H'=H/8$ and $W'=W/8$}
\label{fig:complete_system}
\end{figure*}

\section{Related Work}
\noindent
\paragraph{Single-Person 3D HPE}
\noindent
Single person 3D HPE from a monocular camera has become extremely popular in the last few years. Literature can be classified into three different categories: (i) approaches that first estimate 2D joints and then project them to 3D space, (ii) works that jointly estimate 2D and 3D poses, (iii) methods that learn the 3D pose directly from the RGB image.

The majority of works on single person 3D HPE first compute 2D poses and leverages them to estimate 3D poses, either using off-the-shelf 2D HPE methods \cite{lee2018propagating, hossain2018exploiting, martinez2017simple, mehta2017monocular, bogo2016keep, moreno20173d,chen20173d} or by having a dedicated module in the 3D HPE pipeline \cite{nie2017monocular, pavlakos2018ordinal, lin2017recurrent, yasin2016dual}.

Joint learning of 2D and 3D pose is also shown to be beneficial \cite{mehta2017vnect,dabral2018learning,yang20183d,zhou2017towards,tekin2017learning,omran2018neural,kanazawa2018end,pavlakos2018learning}, often in conjunction with large-scale datasets that only provide 2D pose ground-truth and exploiting anatomical or structure priors.

Finally, recent works estimate 3D pose information directly \cite{sarandi2018synthetic,pavlakos2017coarse,sun2017integral,luvizon20182d,nibali20183d,rhodin2018unsupervised,rhodin2018learning}. Among these, Pavlakos \textit{et al.} \cite{pavlakos2017coarse} were the first to propose a fine discretization of the 3D space around the target by learning a coarse-to-fine prediction scheme in an end to end fashion.


\paragraph{Multi-Person 3D HPE}
\noindent
To the best of our knowledge, very few works tackle multi-person 3D HPE from monocular images. We can categorize them into two classes: top-down and bottom-up approaches.

Top-down methods first identify bounding boxes likely to contain a person using third party detectors and then perform single-person HPE for each person detected. Among them, Rogez \textit{et al.} \cite{LCRNet++} classifies bounding boxes into a set of K-poses. These poses are scored by a classifier and refined using a regressor. The method implicitly reasons using bounding boxes and produces multiple proposals per subject that need to be accumulated and fused. Zanfir \textit{et al.} \cite{Zanfir_2018_CVPR} combine a single person model that incorporates feed-forward initialization and semantic feedback, with additional constraints such as ground plane estimation, mutual volume exclusion, and joint inference. Dabral \textit{et al.} \cite{dabral2018learning}, instead, propose a two-staged approach that first estimates the 2D keypoints in every Region of Interest and then lifts the estimated keypoints to 3D. Finally, Moon \textit{et al.} \cite{moon2019camera} predict absolute 3D human root localization, and root-relative 3D single-person for each person independently. However, these methods heavily rely on the accuracy of the people detector and do not scale well when facing scenes with dozens of people.

In contrast to top-down approaches, bottom-up methods produce multi-person joint locations in a single shot, from which the 3D pose can be inferred even under strong occlusions. Mehta \textit{et al.} \cite{mehta2018single}, predict 2D and 3D poses for all subjects in a single forward pass regardless of the number of people in the scene. They exploit occlusion-robust pose-maps that store 3D coordinates at each joint 2D pixel location. However, their 3D pose read-out strategy strongly depends on the 2D pose output which makes it limited by the accuracy of the 2D module. Their method also struggles to resolve scenes with multiple overlapping people, due to the missing 3D reasoning in their joint-to-person association process. Zanfir \textit{et al.} \cite{zanfir2018deep}, on the other hand, utilize a multi-task deep neural network where the person grouping problem is formulated as an integer program based on learned body part scores parameterized by both 2D and 3D information. Similarly to the latter, our method directly learns a mapping from image features to 3D joint locations, with no need of explicit bounding box detections or 2D proxy poses, while simultaneously being robust to heavy occlusions and multiple overlapping people.

\paragraph{Multi-Person 3D Pose Representation}
\noindent
In a top-down framework, the simplest 3D pose representation can be expressed by a vector of joints. By casting 3D HPE as a coordinate regression task, Rogez \textit{et al.} \cite{LCRNet++} and Zanfir \textit{et al.} \cite{Zanfir_2018_CVPR} indeed utilize x, y, z coordinates of the human joints w.r.t. a known root location. On the other hand, bottom-up approaches require a representation whose coding does not depend on the number of people (e.g. an image map). Among the most recent methods, Mehta \textit{et al.} \cite{mehta2018single} and Zanfir \textit{et al.} \cite{zanfir2018deep} both utilize a pose representation composed by joint-specific feature channels storing the 3D coordinate x, y, or z at the joint/limb 2D pixel location. This representation, however, suffers when multiple overlapping people are present in the scene. In contrast to all these approaches, we adopted the volumetric heatmap representation proposed by Pavlakos \textit{et al.} \cite{pavlakos2017coarse}, overcoming all the limitations that arise when facing a multi-person context.

\section{Proposed Method}
\noindent
The following subsections summarize the key elements of LoCO. Section~\ref{sec:p3dvh} gives a preliminary definition of the chosen volumetric heatmap representation and elaborates on its merits. Section~\ref{sec:autoencoder} illustrates our proposed data mapping which addresses the high dimensional nature of the volumetric heatmaps by producing a compact and more tractable representation. Next, in Section~\ref{sec:code_predictor}, we describe how our strategy can be easily exploited to effectively tackle the problem of multi-person 3D HPE in a single-shot bottom-up fashion. Finally, Section~\ref{sec:refiner} illustrates our simple refining approach that prevents poses from being implausible.

\subsection{Volumetric Heatmaps}
\label{sec:p3dvh}
\noindent
By considering a voxelization of the RGB-D volumetric space \cite{dai20183dmv,pavlakos2017coarse}, we refer as a volumetric heatmap, ${\mathfrak{h}}$, the 3D confidence map with size $D \times H \times W$, where $D$ represents the depth dimension (appropriately quantized), while $H$ and $W$ represent the height and width of the image plane respectively. Given the body joint $j$ with pseudo-3D coordinates $\mathbf{u}_j = \left(u_{1,j}, u_{2,j}, u_{3,j} \right)$, where $u_{1,j} \in \{1, ..., D\}$ is the quantized distance of joint $j$ from the camera, and $u_{2,j} \in \{1, ..., H\}$ and $u_{3,j} \in \{1, ..., W\}$ are respectively the row and column indexes of its pixel on the image plane, the value of ${\mathfrak{h_j}}$ at a generic location $u$ is obtained by centering a fixed variance Gaussian in $u_j$:

\begin{equation}\label{eq:vol_hmap}
    {\mathfrak{h_j}}(\mathbf{u}) = 
    e^{-\frac{\left\|  \mathbf{u}-\mathbf{u}_j \right \|  ^2}{\sigma^2}}
\end{equation}

In a multi-person context, in the same image we can simultaneously have several joints of the same kind (e.g. ``left ankle"), one for each of the $K$ different people in the image. 
In this case we aggregate those $K$ volumetric heatmaps ${\mathfrak{h}_j}^{(k)}$, into a single heatmap ${\mathfrak{h}_j}$ with a max operation:

\begin{equation}\label{eq:max_aggregation}
    {\mathfrak{h}_j}(\mathbf{u}) = \textup{max}_{k}\{{\mathfrak{h}_j}^{(k)}(\mathbf{u})\}
\end{equation}

Finally, considering $N$ different types of joint and $K$ people, we have a set of $N$ volumetric heatmaps (each associated with a joint type),  ${\mathfrak{H}} = \{{\mathfrak{h}}_j, j=1,...,N\}$, resulting from the aggregation of the individual heatmaps of the $K$ people in the scene. Note that, given pseudo-3D coordinates $\mathbf{u} = \left(u_1, u_2, u_3 \right)$ and the camera intrinsic parameters, i.e. focal length $f=\left( f_x, f_y\right)$ and principal point $\left(c_x, c_y \right)$, the corresponding 3D coordinates $\mathbf{x}=\left(x, y, z \right)$ in the camera reference system can be retrieved by directly applying the equations of the pinhole camera model.

The benefit of choosing a volumetric heatmap representation over a direct 3D coordinate regression is that it casts the highly non-linear problem to a more tractable configuration of prediction in a discretized space. In fact, joint predictions do not estimate a unique location but rather a per voxel confidence, which makes it easier for a network to learn the target function \cite{pavlakos2017coarse}. In the context of 2D HPE, the benefits of predicting confidences for each pixel instead of image coordinates are well known~\cite{pfister2015flowing,tompson2014joint}. Moreover, in a multi-person environment, directly regressing the joint coordinates is unfeasible when the number of people is not known a priori, making volumetric heatmaps a natural choice for tackling bottom-up multi-person 3D HPE.

The major disadvantage of this representation is that it is memory and computational demanding, requiring some compromise during implementation that limits its full potential. Some of those compromises consist in utilizing low resolution heatmaps that introduce quantization errors or complex training strategies that involve coarse-to-fine predictions through iterative refining of network output \cite{pavlakos2017coarse}.
\begin{table}[t]
\centering
\label{tab:autoencoder}
\begin{tabular}{lcccc}
\hline
block & layer & in ch.   & out ch.  & stride\\
\hline
\\[-0.7em]
\multirow{3}{*}{\textit{e}-c2d}   
& Conv2D + ReLU  &  $D$      & $D/d_1$       & $s_1$        \\
& Conv2D + ReLU  &  $D/d_1$      & $D/d_2$    & $s_2$      \\
& Conv2D + ReLU  &  $D/d_2$    & $D/d_3$     & $s_2$        \\[0.7em]
\multirow{2}{*}{\textit{e}-c3d}      
& Conv3D + ReLU  &  $N$      & 4        & 1          \\
& Conv3D + ReLU  &  4        & 1         & 1           \\ 
\hline
\end{tabular}
\caption{Structure of the encoder part of the Volumetric Heatmap Autoencoder (VHA). The decoder is not shown as it is perfectly mirrored to the encoder. \textit{VHAv1}: $(d_1, d_2, d_3)=(1,2,2)$ and $(s_1, s_2, s_3)=(1,2,1)$; for \textit{VHAv2}: $(d_1, d_2, d_3)=(2,4,4)$ and $(s_1, s_2, s_3)=(2,2,1)$; \textit{VHAv3}: $(d_1, d_2, d_3)=(2,4,8)$ and $(s_1, s_2, s_3)=(2,2,2)$}
\label{tab:architecture}
\end{table}

\subsection{Volumetric Heatmap Autoencoder}
\label{sec:autoencoder}
\noindent
\label{sec:heatmap_autoencoder}
To overcome the aforementioned limitations without introducing quantization errors or training complexity, we propose to map volumetric heatmaps to a more tractable representation. Inspired by \cite{luo2018fast}, we propose a multiple branches Volumetric Heatmap Autoencoder (VHA) that takes a set of $N$ volumetric heatmaps ${\mathfrak{H}}$ as input. 
At first, the volumetric heatmaps $\{\mathfrak{h}_1, ..., \mathfrak{h}_N\}$ are processed independently with a 2D convolutional block (\textit{e}-c2d) in which the kernel does not move along the $D$ dimension. In order to capture the mutual influence between joints locations, the obtained maps are then stacked along a fourth dimension and processed by a subsequent set of 3D convolutions (\textit{e}-c3d). The resulting encoded representation, $e(\mathbf{{\mathfrak{H}}})$ is finally decoded by its mirrored architecture $d\left( e \left(\mathbf{{\mathfrak{H}}} \right)\right) = {\tilde{\mathfrak{H}}}$. The general structure of the model is outlined in Fig.~\ref{fig:complete_system} top.


The goal of the VHA is therefore to learn a compressed representation of the input volumetric heatmaps that preserve their information content, which results in the preservation of the position of the Gaussian peaks of the various joints in the original maps. For the purpose, we maximize the F1-score, $\textup{F1}\left( Q_{{\mathfrak{H}}}, Q_{\tilde{{\mathfrak{H}}}} \right)$, between the set of ground truth peaks ($Q_{{\mathfrak{H}}}$) and the set of the decoded maps ($Q_{\tilde{{\mathfrak{H}}}}$). We define the set of peaks as follows:

\begin{equation}
\label{eq:peaks}
    Q_{{\mathfrak{H}}}=\bigcup_{n=1,...,N} \left \{ \mathbf{u}:\mathfrak{h}_n\left ( \mathbf{u} \right )>\mathbf{u}' \, \, \, \forall  \mathbf{u}' \in  \mathfrak{N}_\mathbf{\bar{u}} \right \}
\end{equation}

where $\mathfrak{N}_\mathbf{\bar{u}}$ is the 6-connected neighborhood of $\mathbf{\bar{u}}$, i.e. the set of coordinates $\mathfrak{N}_\mathbf{\bar{u}} = \left \{ \mathbf{u}: \left \| \mathbf{u}-\mathbf{\bar{u}} \right \| = 1\right \}$ at unit distance from $\mathbf{\bar{u}}$. Since the procedure for extracting the coordinate sets from the volumetric heatmaps is not differentiable, the former objective cannot be directly optimized as a loss component for training the VHA.  To address this issue, we propose to use mean squared error (MSE) loss between $\mathfrak{H}$ and $\tilde{\mathfrak{H}}$ as training loss.

Note that our proposed mapping purposely reduces the volumetric heatmap's fourth dimension, making its shape coherent with the output of 2D convolutions and thus exploitable by regular CNN backbones. Additional architecture details can be found in the supplementary material.

\begin{table*}
\centering
\begin{tabular}{cclccclccclccc}
 & & & 
\multicolumn{3}{c}{F1 on JTA} &  & 
\multicolumn{3}{c}{F1 on Panoptic} &  & 
\multicolumn{3}{c}{F1 on Human3.6m} \\ 
\cline{4-6} \cline{8-10} \cline{12-14}  \\[-0.9em]
model & bottleneck size &  & 
@0vx  & @1vx  & @2vx  &  & 
@0vx  & @1vx  & @2vx  &  & 
@0vx  & @1vx  & @2vx  \\ 
\hline
\\[-0.8em]

VHA$^{(1)}$ & $ \frac{D}{2} \times \frac{H'}{2}  \times \frac{W'}{2} $ &  
& 97.1   & 98.4   & 98.5   &  
& -      & -      & -      &  
& -      & -      & -      \\[0.2em]

VHA$^{(2)}$ & $ \frac{D}{4} \times \frac{H'}{4}  \times \frac{W'}{4} $ &  
& 92.5   & 97.0   & 97.1    &  
& 97.1   & 98.6   & 98.9    &
& 100.0   & 100.0  & 100.0   \\[0.2em]

VHA$^{(3)}$ & $ \frac{D}{8} \times \frac{H'}{8}  \times \frac{W'}{8} $ &  
& 56.5   & 90.3    & 92.9    &  
& 91.9   & 98.7    & 99.6    &  
& 99.7   & 100.0   & 100.0   \\[0.2em]

\hline
\end{tabular}
\caption{VHA bottleneck/code size and performances on the JTA, Panoptic and Human3.6m (protocol P2) test set in terms of F1 score at different thresholds @0, @1, and @2 voxel(s); @\textit{t} indicates that a predicted joint is considered “true positive” if the distance from the corresponding ground truth joint is less than \textit{t}}
\label{tab:vha_exp_new}
\end{table*}

\subsection{Code Predictor and Body Joints Association}
\label{sec:code_predictor}
\noindent
The input of the Code Predictor is represented by a RGB image, $\mathbf{I}$, while its output, $f\left(\mathbf{I}\right)$, aims to predict the codes obtained with the VHA, Fig.~\ref{fig:complete_system}. The architecture, Fig.~\ref{fig:complete_system} bottom, is inspired by \cite{simple_baselines} thus composed by a pre-trained feature extractor (convolutional part of Inception v3 \cite{inception}), and a fully convolutional block (\textit{f}-c2d) composed of four convolutions. We trained the Code Predictor by minimizing the MSE loss between $f\left(\mathbf{I}\right)$ and $e\left(\mathfrak{H}\right)$, where $\mathfrak{H}$ is the volumetric heatmap associated with the image $\mathbf{I}$.

At inference time, the pseudo-3D coordinates of the body joints are obtained from the decoded volumetric heatmap $\mathfrak{\tilde{H}} = d(f\left(\mathbf{I}\right))$ through a local maxima search. Eventually, if camera parameters are available, the pinhole camera equations recover the true three-dimensional coordinates of the detected joints. Additional details in the supplementary material.

As in almost all recent 2D HPE bottom-up approaches \cite{cao2017realtime,Fieraru_2018_CVPR_Workshops,Chen_2018_CVPR} (i.e. methods which does not require a people detection step) detected joints have to be linked together to obtain people skeletal representations. In a single person context, joint association is trivial. On the other hand, in a multi-person environment, linking joints is significantly more challenging. For the purpose, we rely on a simple distance-based heuristic where, starting from detected heads (i.e. the joint with the highest confidence), we connect the remaining $(N-1)$ joints by selecting the closest ones in terms of 3D Euclidean distance. Associations are further refined by rejecting those that violates anatomical constraints (e.g. length of a limb greater than a certain threshold). Despite its simplicity, this approach is particularity effective when 3D coordinates of body joints are available, especially in surveillance scenarios where proxemics dynamics often regulate the spatial relationships between different individuals. Additional details are reported in the supplementary material.

\begin{table*}[h!]
\small
\centering
\begin{tabular}{lccccccccccccccc}
\multicolumn{1}{c}{}      &  & PR    & RE    & F1   &  & PR    & RE    & F1   &  & PR    & RE    & F1   
\\ \cline{3-5} \cline{7-9} \cline{11-13} \\[-1em]
\multicolumn{1}{c}{} 
&  & \multicolumn{3}{c}{@0.4 m}                                            
&  & \multicolumn{3}{c}{@0.8 m}       
&  & \multicolumn{3}{c}{@1.2 m} \\  
\hline

\\[-0.9em]
Location Maps \cite{mehta2018single,mehta2017vnect}
&  &  5.80 &  5.33 &  5.42
                      &  & 24.06 & 21.65 & 22.29
                      &  & 41.43 & 36.96 & 38.26
                      \\
Location Maps \cite{mehta2018single,mehta2017vnect} + ref. 
&  &  5.82 &  5.89 &  5.77
                      &  & 23.28 & 23.51 & 23.08
                      &  & 38.85 & 39.17 & 38.49
                      \\
\cite{redmon2018yolov3} + \cite{martinez2017simple}          
                      &  & \textbf{75.88} & 28.36 & 39.14
                      &  & \textbf{92.85} & 34.17 & 47.38
                      &  & \textbf{96.33} & 35.33 & 49.03
                      \\
Uncompr. Volumetric Heatmaps
                      &  & 25.37 & 24.40 & 24.47 
                      &  & 45.40 & 43.11 & 43.51 
                      &  & 55.55 & 52.44 & 53.08 
                      \\
\hline
\\[-0.8em]
LoCO$^{(1)}$         & & 48.10 & 42.73 & 44.76
                    & & 65.63 & 58.58 & 61.24
                    & & 72.44 & 64.84 & 67.70
                      \\
LoCO$^{(1)}+$.  & & 49.37 & 43.45 & 45.73
                    & & 66.87 & 59.02 & 62.02
                    & & 73.54 & 65.07 & 68.29

                      \\
LoCO$^{(2)}$        & & 54.76 & 46.94 & 50.13
                    & & 70.67 & 60.48 & 64.62
                    & & 77.00 & 65.92 & 70.40
                      \\
LoCO$^{(2)}+$. & & 55.37 & \textbf{47.84} & \textbf{50.82}
                & & 70.63 & \textbf{60.94} & \textbf{64.76}
                & & 76.81 & \textbf{66.31} & \textbf{70.44}

                      \\[0.2em]
LoCO$^{(3)}$        & & 48.18 & 41.97 & 44.49
                    & & 66.96 & 58.22 & 61.77
                    & & 74.43 & 64.71 & 68.65

                      \\
LoCO$^{(3)}+$.  & & 49.15 & 42.84 & 45.36
                    & & 67.16 & 58.45 & 61.92
                    & & 74.39 & 64.76 & 68.57
                      \\
\hline
\\[-0.9em]
GT Location Maps \cite{mehta2018single,mehta2017vnect}
                       &  & 76.07 & 64.83 & 69.59 
                       &  & 76.07 & 64.83 & 69.59 
                       &  & 76.07 & 64.83 & 69.59 
                       \\
GT Volumetric Heatmaps 
                       &  & 99.96 & 99.96 & 99.96 
                       &  & 99.99 & 99.99 & 99.99 
                       &  & 99.99 & 99.99 & 99.99 
                       \\[0.2em]
\hline
\end{tabular}
\caption{Comparison of our LoCO approach with other strong baselines and competitors on the JTA test set. In PR (precision), RE (recall) and F1, @$t$ indicates that a predicted joint is considered ``true positive" if the distance from the corresponding ground truth joint is less than $t$. Last two rows contain the upper bounds obtained using the ground truth location maps and volumetric heatmaps respectively}
\label{tab:cp_exp_jta}
\end{table*}


\subsection{Pose Refiner}
\label{sec:refiner}
\noindent
The predicted 3D poses are subsequently refined by a MLP network trained to account for miss-detections and location errors.
The objective of the Pose Refiner is indeed to make sure that the detected poses are complete (i.e. all the $N$ joints are always present).
To better understand how the Pose Refiner works, we define the concept of \emph{3D poses} and \emph{root-relative poses}. Given a person $k$, its 3D pose is the set $\mathbf{p}^{(k)}= \left \{\mathbf{x}_n^{(k)}, \,\, n=1,...,N \right \}$ of the 3D coordinates of its $N$ joints. The corresponding root-relative pose is then given by:

\begin{equation}
\label{eq:fountain}
\mathbf{p_{rr}}^{(k)}= \left \{
    \frac{\mathbf{x}_n^{(k)} -  \mathbf{x}_1^{(k)}}{l_n},\,\, n=2,...,N  
\right \}
\end{equation}

where $\mathbf{x}_1$ are the 3D coordinates of the root joint (``head-top'' in our experiments) and $l_n$ is a normalization constant computed on the training set as the maximum length of the vector that points from the root joint to any other joint of the same person.

The Pose Refiner is hence trained with MSE loss taking as input the root-relative version of the 3D poses with randomly removed joints, and an additional Gaussian noise applied to the coordinates. Given the 3D position of the root joint and the refined poses, it is straightforward to re-obtain the corresponding 3D poses by using Eq.~\eqref{eq:fountain}.

\section{Experiments}
\label{sec:experiments}
\noindent
A series of experiments have been conducted on two multi-person datasets, namely JTA \cite{fabbri2018learning} and CMU Panoptic \cite{panoptic_1, panoptic_2, panoptic_3}, as well as one well established single-person benchmark: Human3.6m \cite{h36m_pami}.

JTA is a large synthetic dataset for multi-person HPE and tracking in urban scenarios. It is composed of 512 Full HD videos, 30s long, each containing an average of 20 people per frame. Due to its recent publication date, this dataset does not have a public leaderboard and it is not mentioned in other comparable HPE works. Despite this limitation, we believe it is crucial to test LoCO on JTA because it is much more complex and challenging than older benchmarks.

CMU Panoptic is another large dataset containing both single-person and multi-person sequences for a total of 65 sequences (5.5 hours of video). It is less challenging than JTA as the number of people per frame is much more limited, but it is currently the largest real-world multi-person dataset with 3D annotations.

To further demonstrate the generalization capabilities of LoCO, we also provide a direct comparison with other HPE approaches on the single person task. Without any modification to the multi-person pipeline, we achieve state of the art results on the popular Human3.6m dataset.

For each dataset we also show the upper bound obtained by using the GT volumetric heatmaps in order to highlight the strengths of this data representation. In all the following tables, we will indicate with LoCO$^{(n)}$ our complete HPE pipeline, composed of the Code Predictor, the decoder of VHA$^{(n)}$ and the subsequent post-processing. LoCO$^{(n)}+$ is the same system with the addition of the Pose Refiner.

For all the experiments in the paper we utilized Adam optimizer with learning rate $10^{-4}$. We employed batch size 1 when training the VHA and batch size 8 when training the Code Predictor. We employed Inception v3 \cite{inception} as backbone for the Code Predictor, which is followed by 3 convolutions with ReLU activation having kernel size 4 and with 1024, 512 and 256 channels respectively. A last $1\times1$ convolution is performed to match the compressed volumetric heatmap's number of channels. Additional training details in the supplementary material.

\subsection{Compression Levels}
\label{subsec:vha_exp}
\noindent
In order to understand how different code sizes in the VHA affects the performance of our Code Predictor network, multiple VHA versions have been tested. Specifically, we designed three VHA versions with decreasing bottleneck sizes. Each version has been trained on JTA first and then finetuned on CMU Panoptic and Human3.6m. VHA's architecture details are depicted in Tab.~\ref{tab:architecture} for every version. 

As shown in Tab.~\ref{tab:vha_exp_new}, as the bottleneck size decreases, there is a corresponding decrease in the F1-score. Intuitively, the more we compress, the less information is being preserved. VHA$^{(1)}$ is only considered when using JTA, as VHA$^{(2)}$ and VHA$^{(3)}$ already obtain an almost lossless compression on Panoptic and Human3.6m, due to their smaller number of people in the scene. 

All the experiments has been conducted considering a 14 joints volumetric heatmap representation of shape $14 \times D \times H' \times W'$, where $H'$ and $W'$ are height and width downsampled by a factor of 8, while $D$ has been fixed to 316 bins. Note that the real-world depth grid covered by our representation is a uniform discretization in $[0, 100]$m for JTA, $[0, 7]$m for Panoptic and $[1.8, 8.1]$m for Human3.6m. Thus, every bin has a depth size of approximately $0.32$m for JTA and $0.02$m for Panoptic and Human3.6m.

\begin{figure*}[t]
\begin{center}
\includegraphics[width=0.24\textwidth]{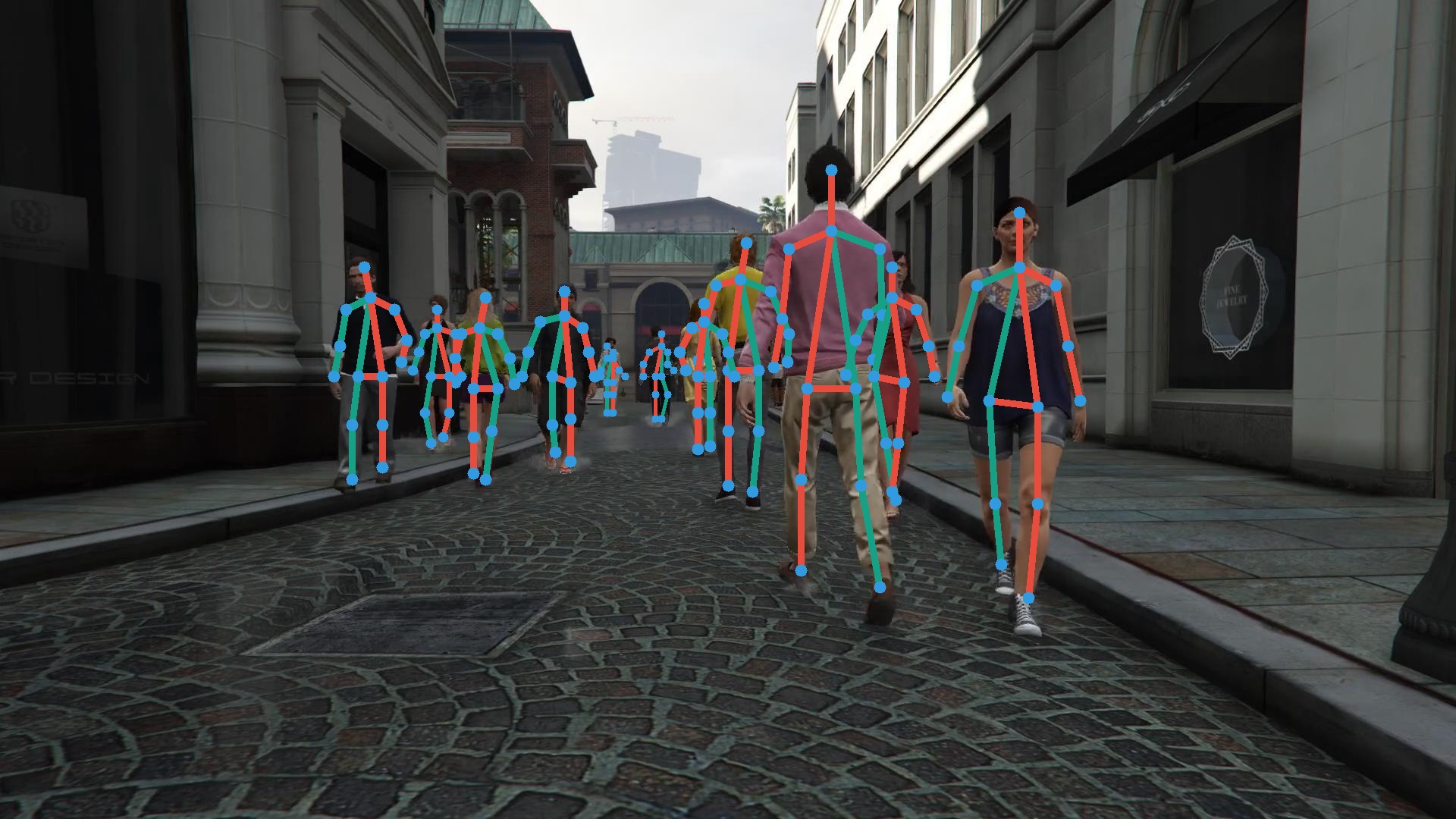}
\includegraphics[width=0.24\textwidth]{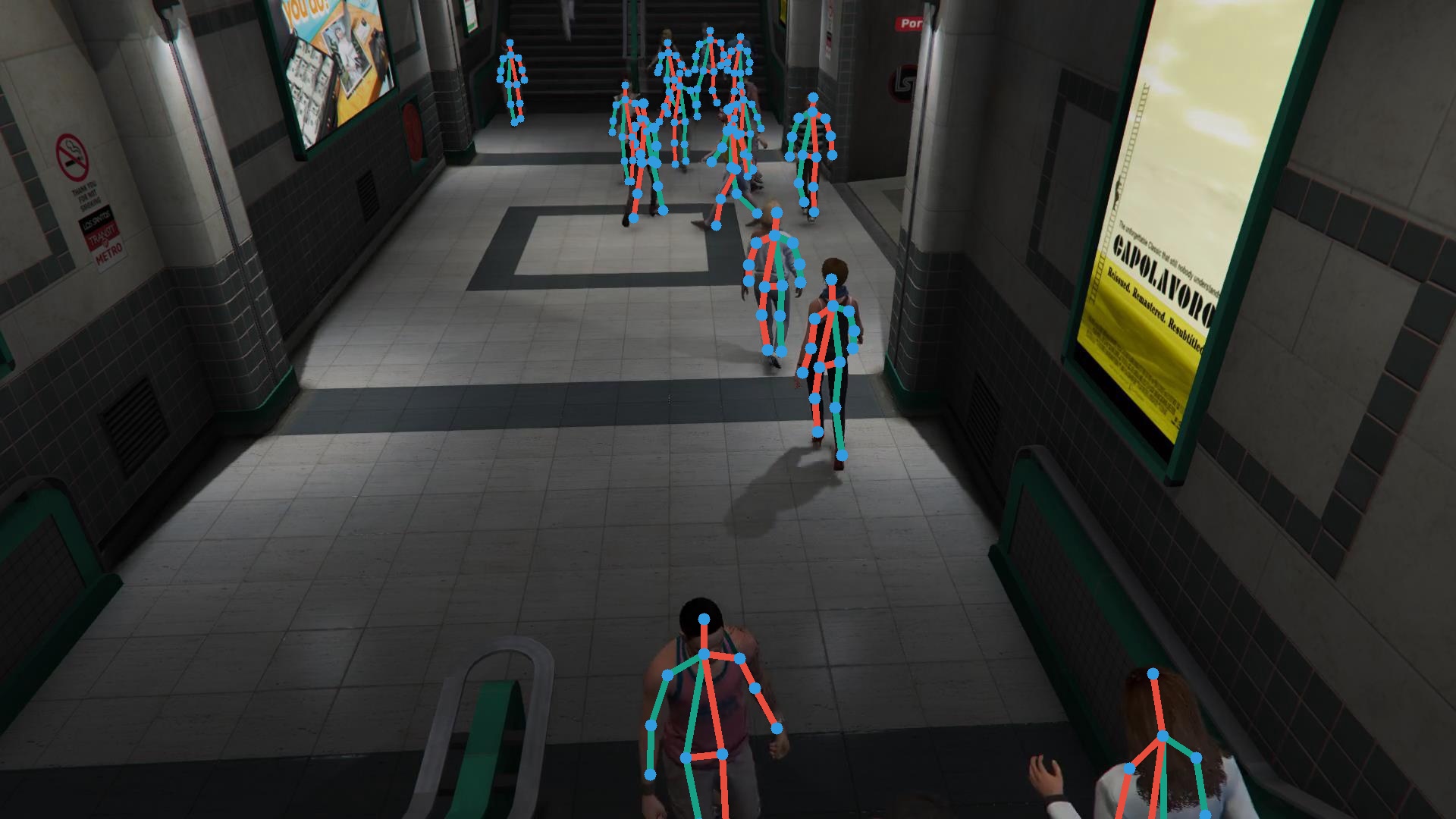}
\includegraphics[width=0.24\textwidth]{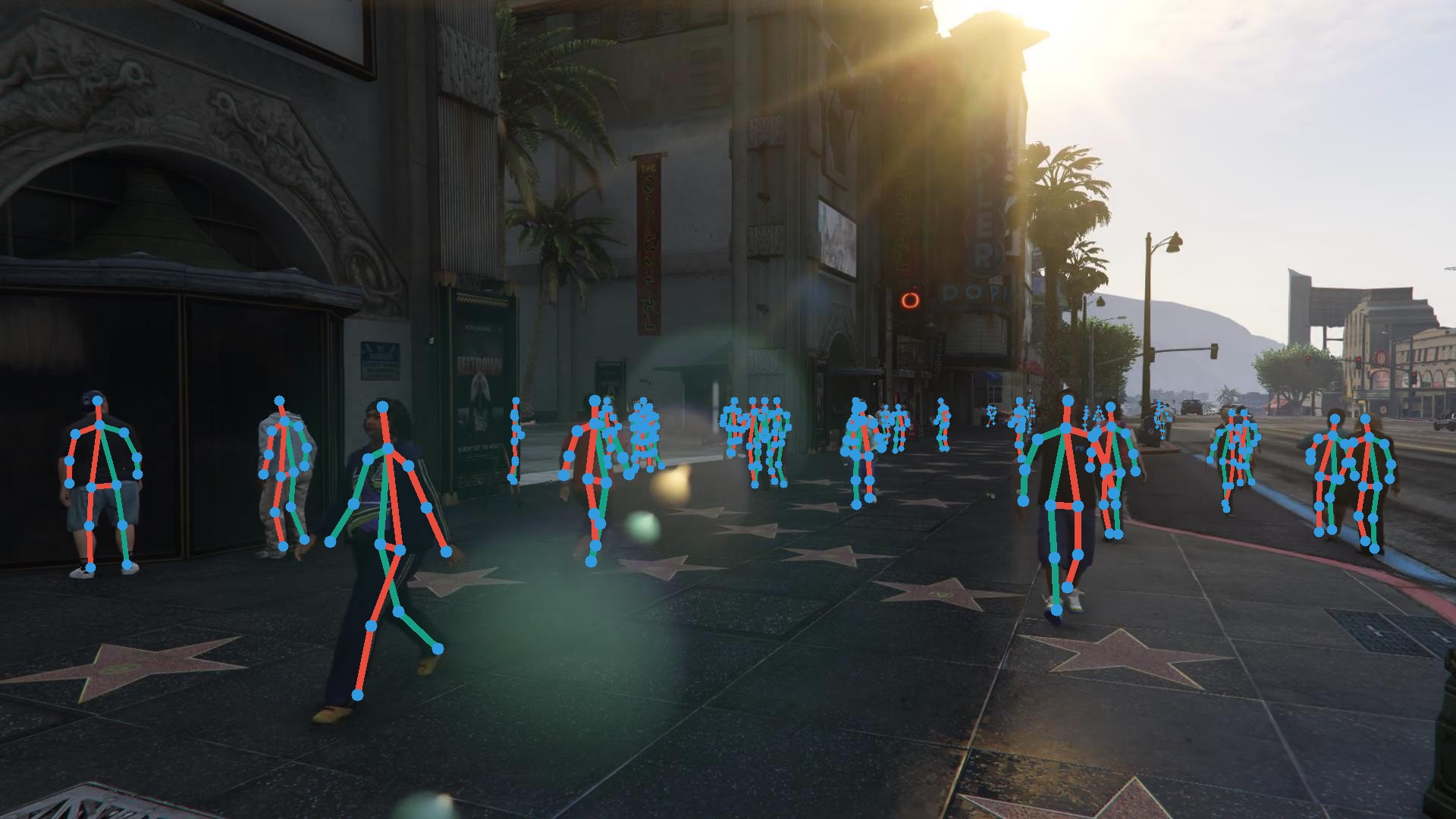}
\includegraphics[width=0.24\textwidth]{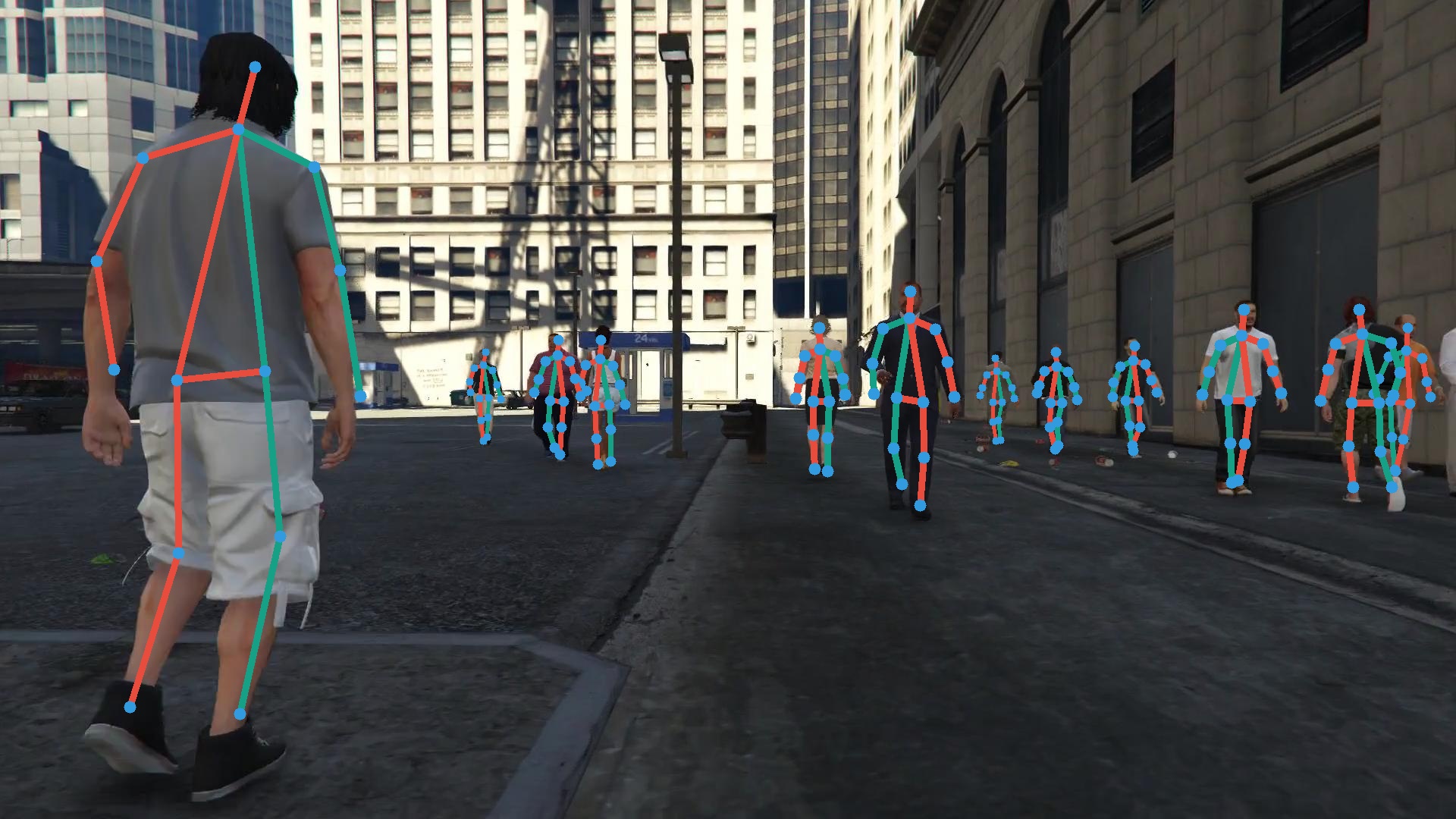}
\\[1px]
\includegraphics[width=0.24\textwidth]{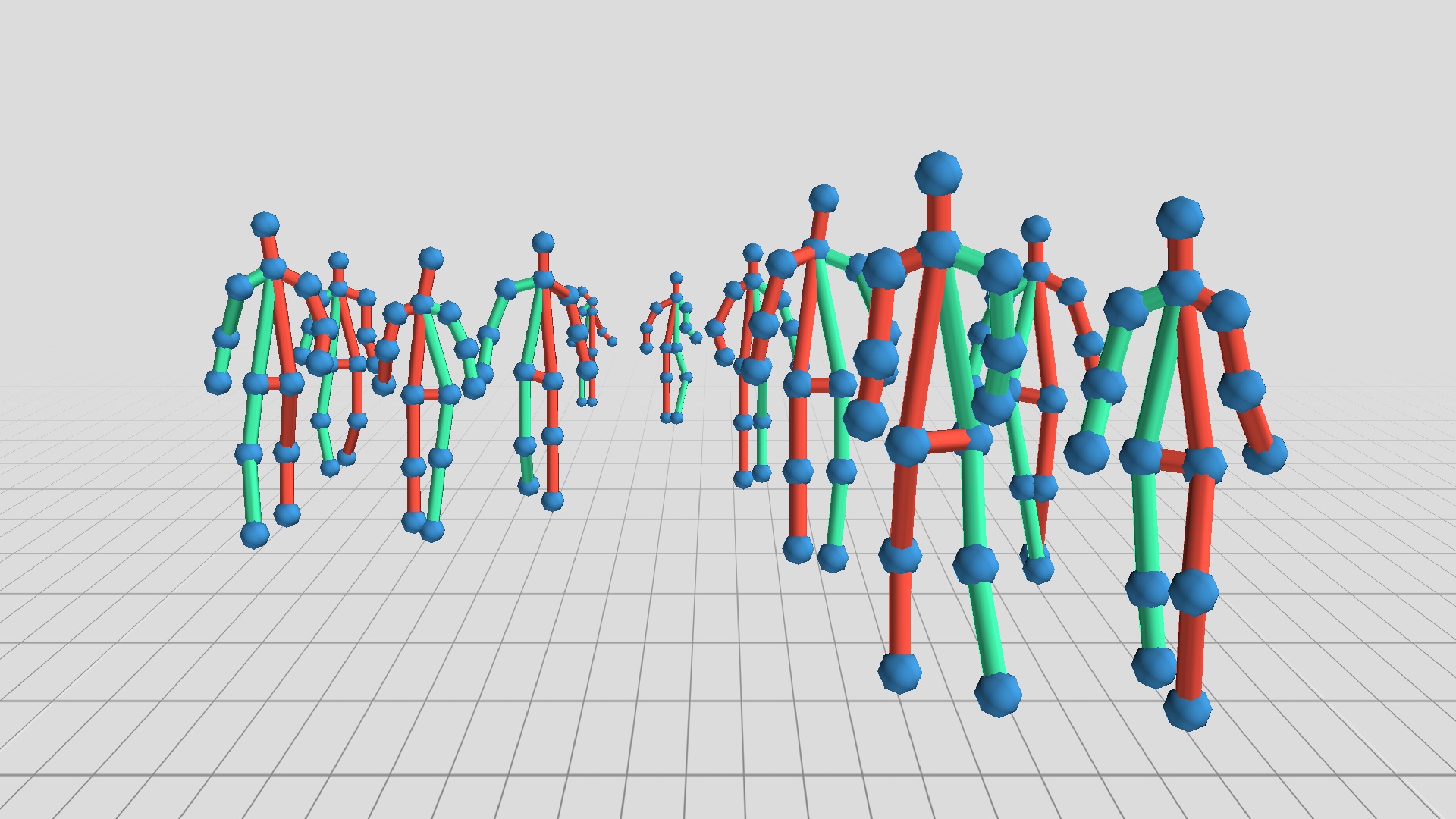}
\includegraphics[width=0.24\textwidth]{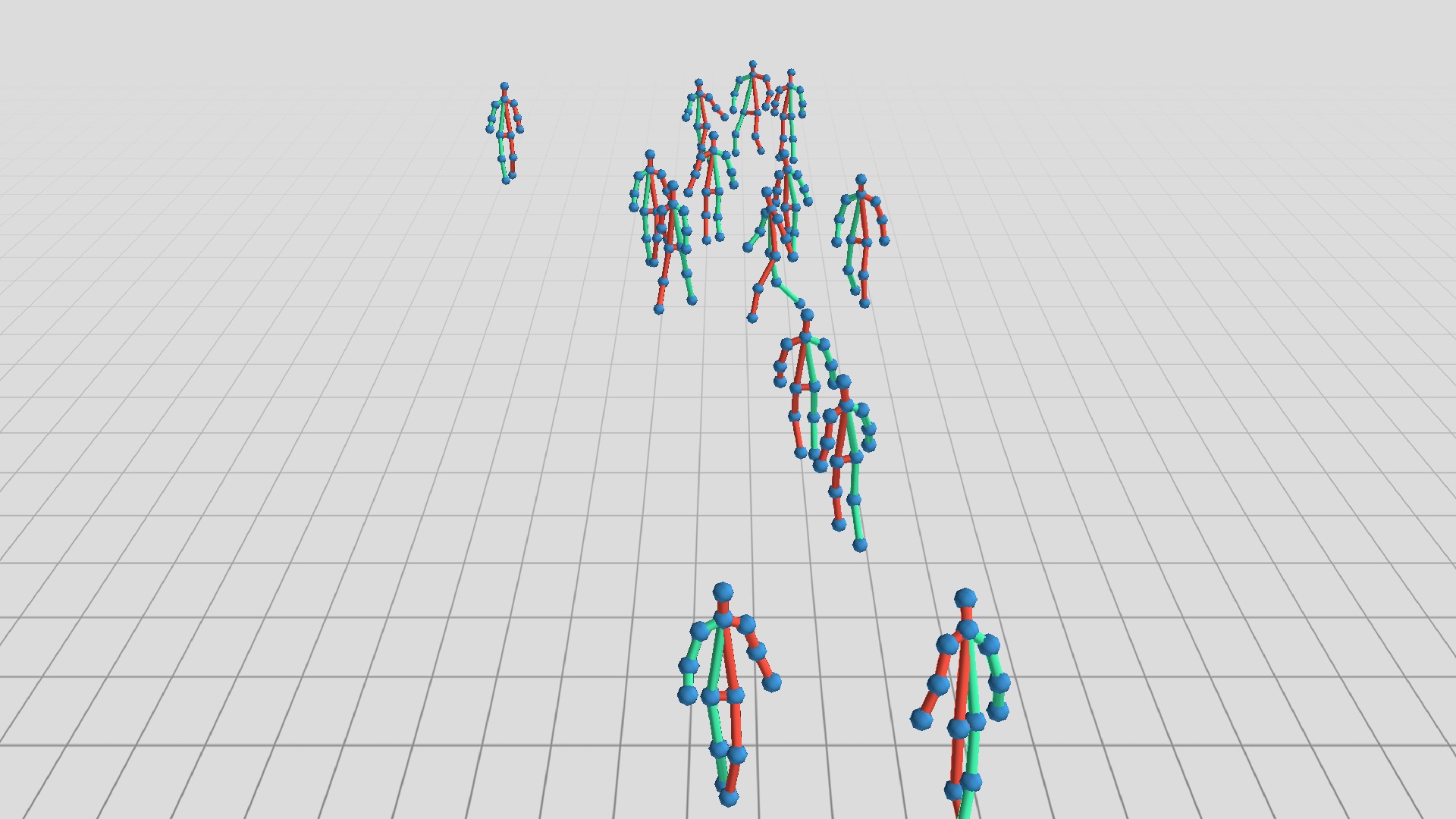}
\includegraphics[width=0.24\textwidth]{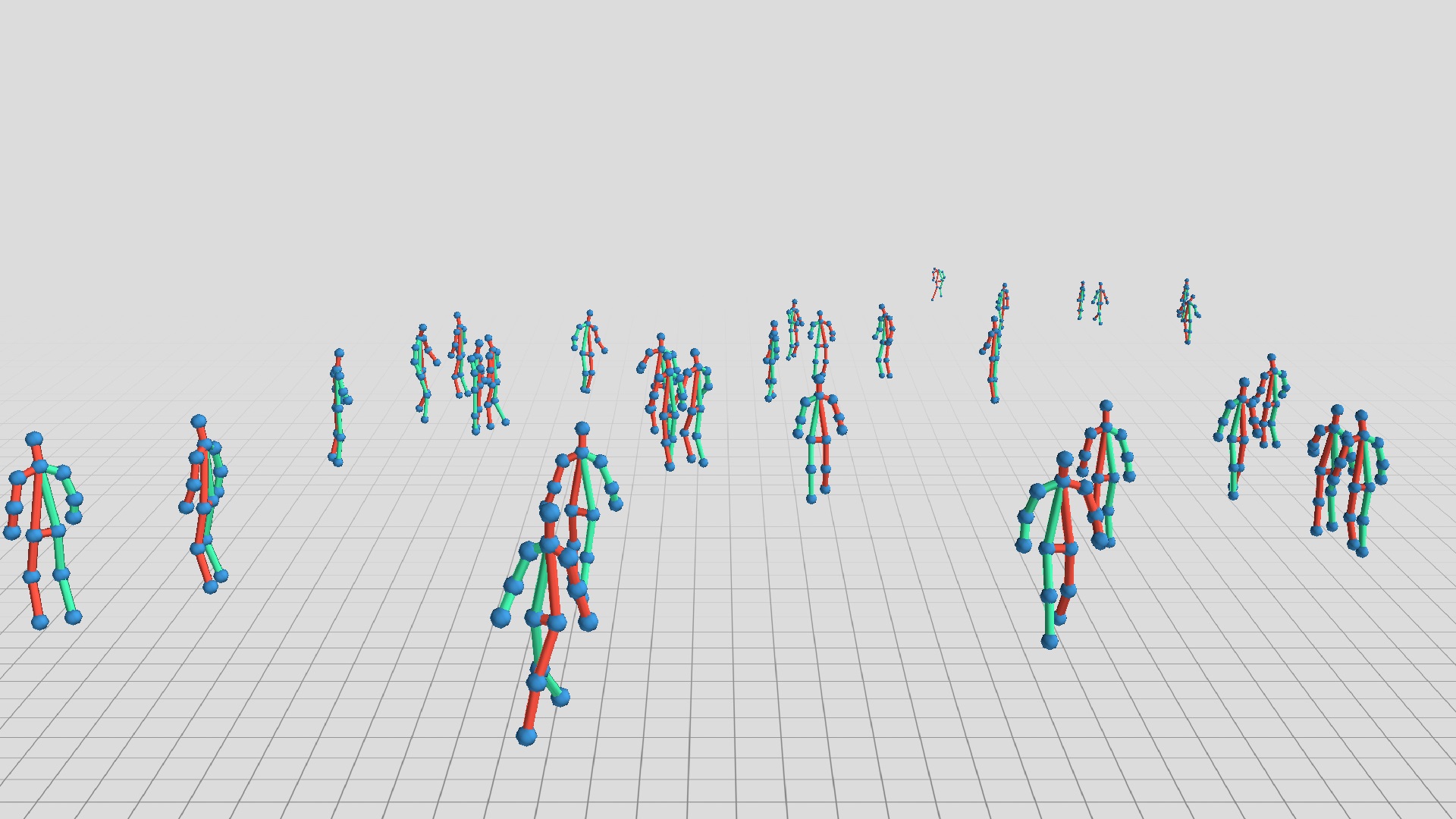}
\includegraphics[width=0.24\textwidth]{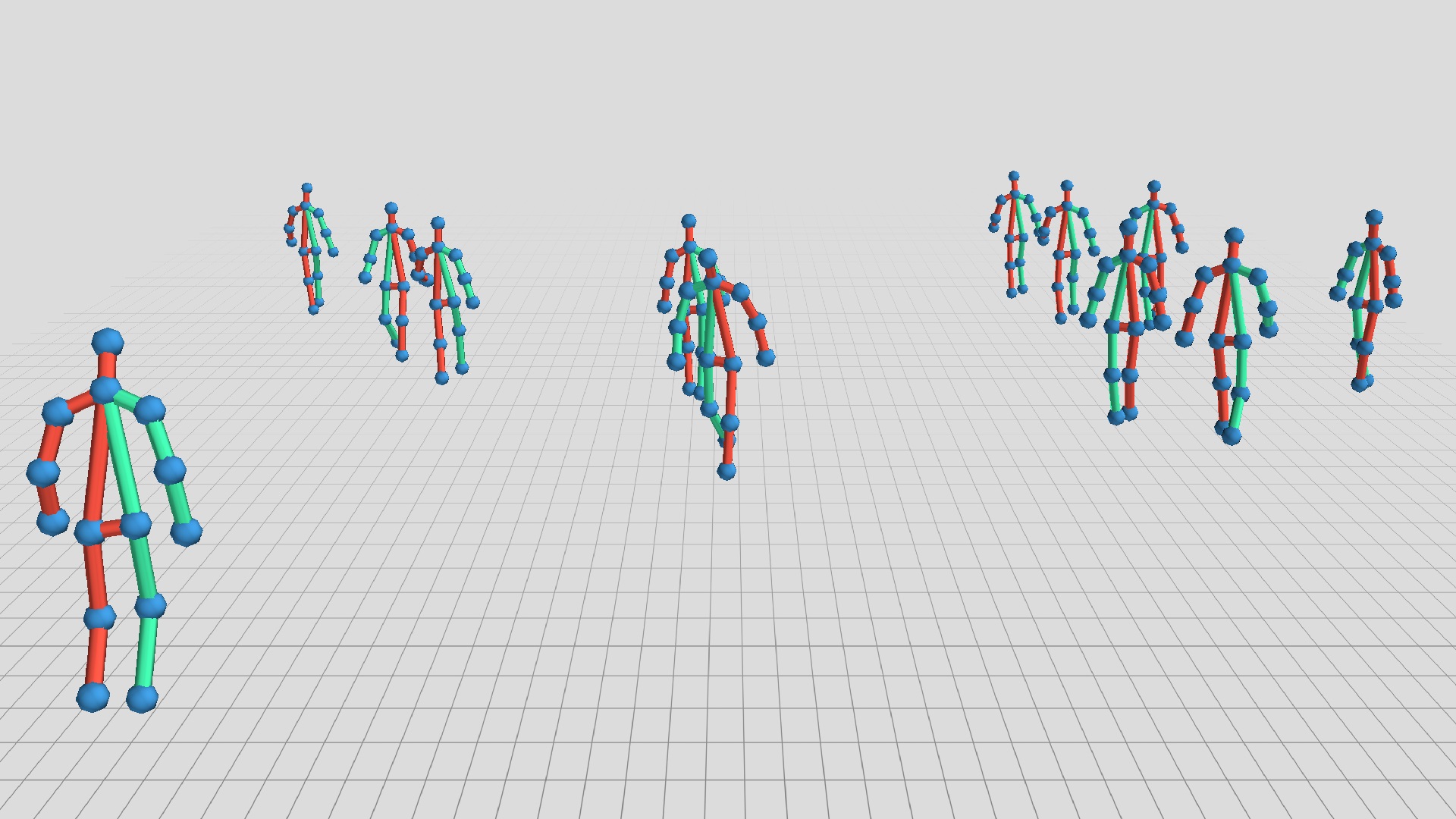}
\\[1px]
\includegraphics[width=0.24\textwidth]{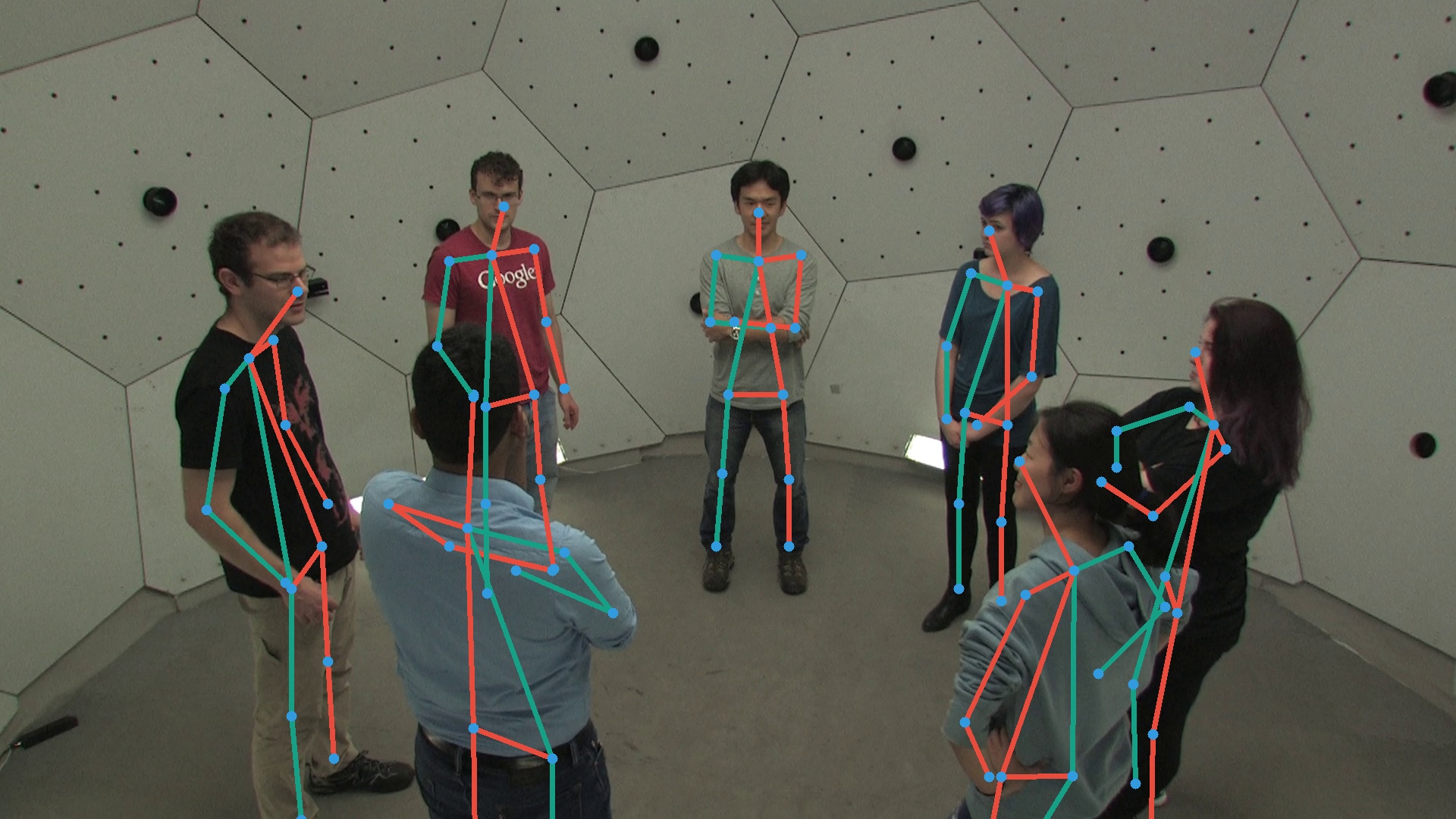}
\includegraphics[width=0.24\textwidth]{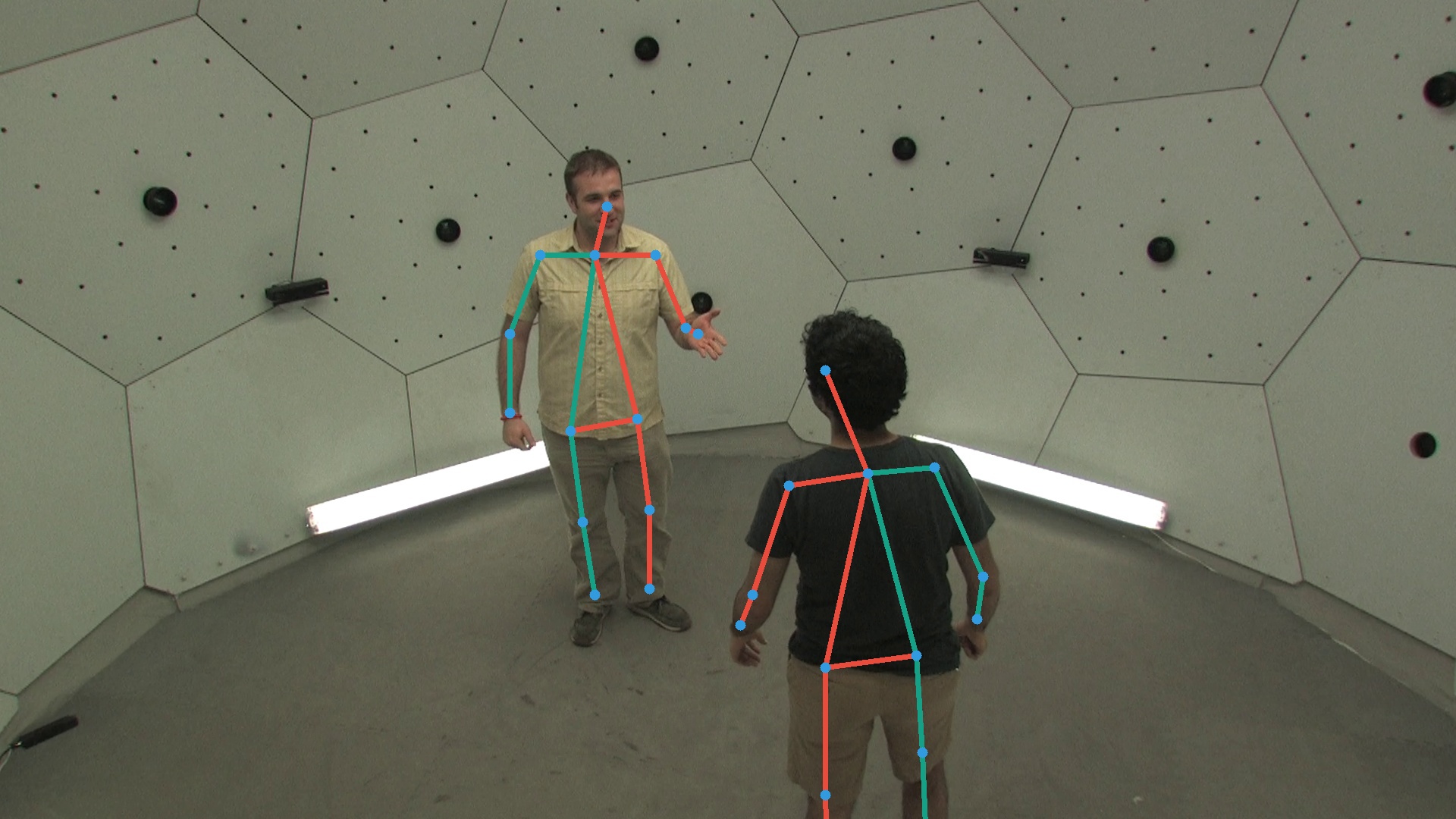}
\includegraphics[width=0.24\textwidth]{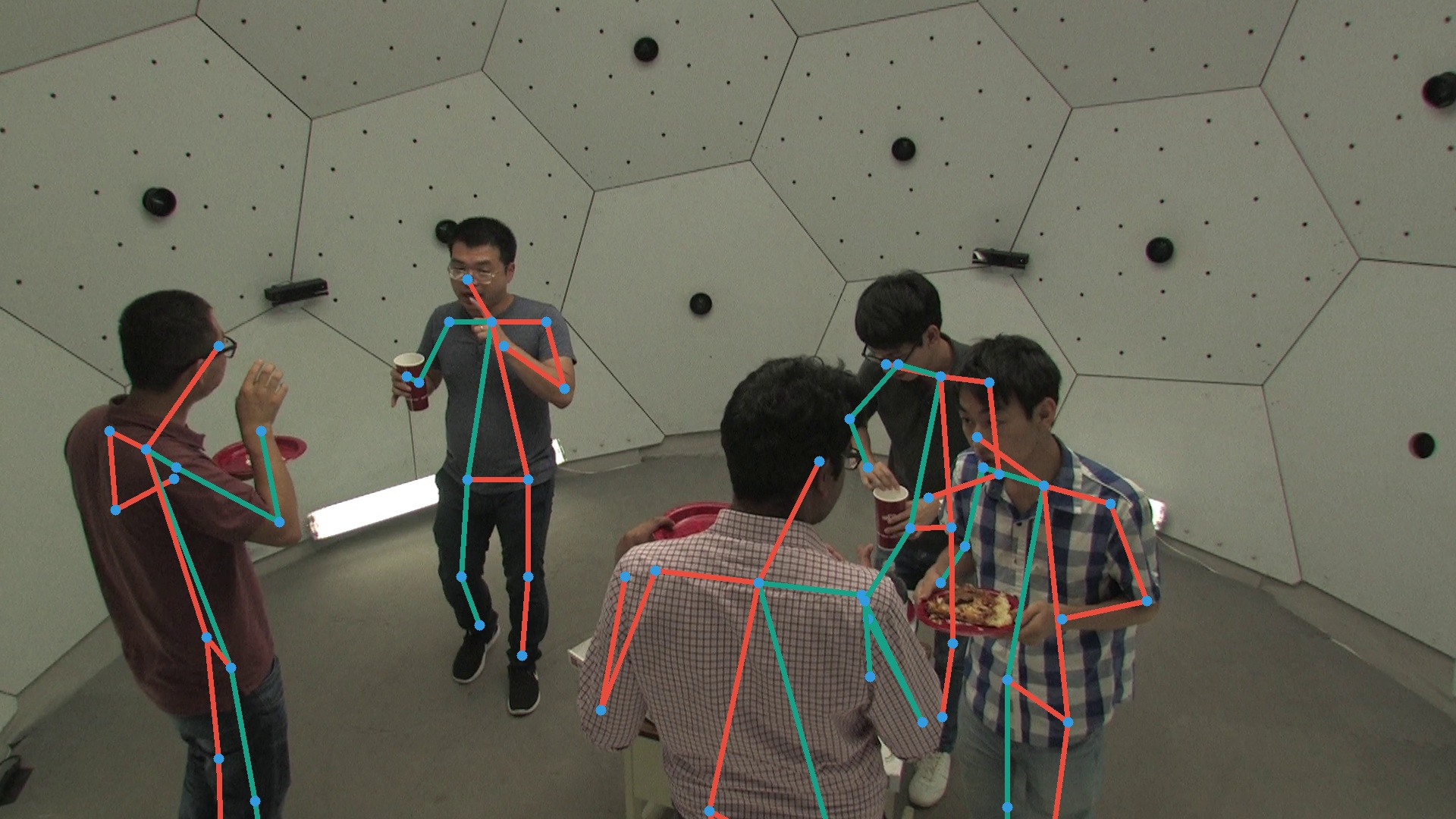}
\includegraphics[width=0.24\textwidth]{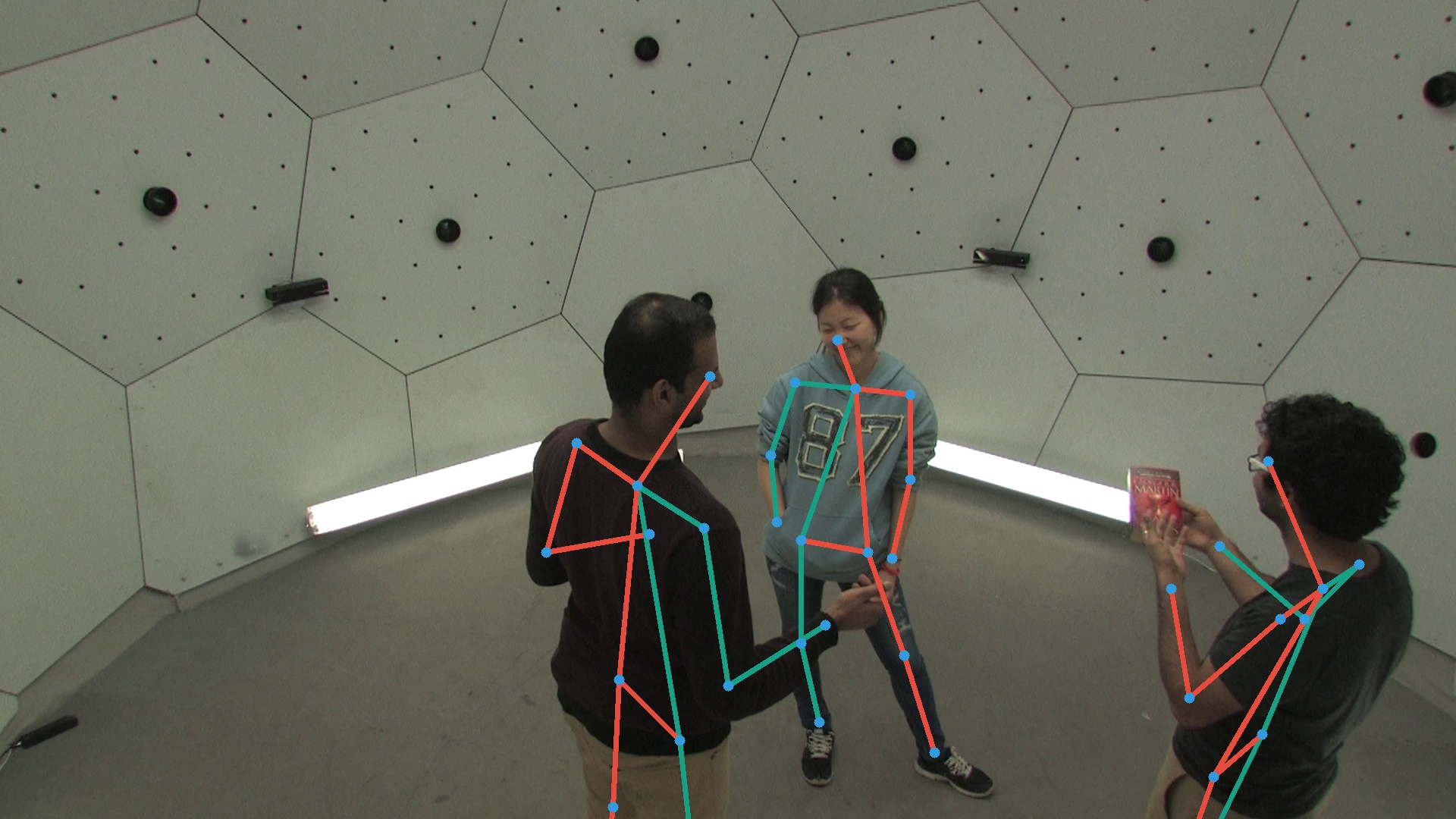}
\\[1px]
\includegraphics[width=0.24\textwidth]{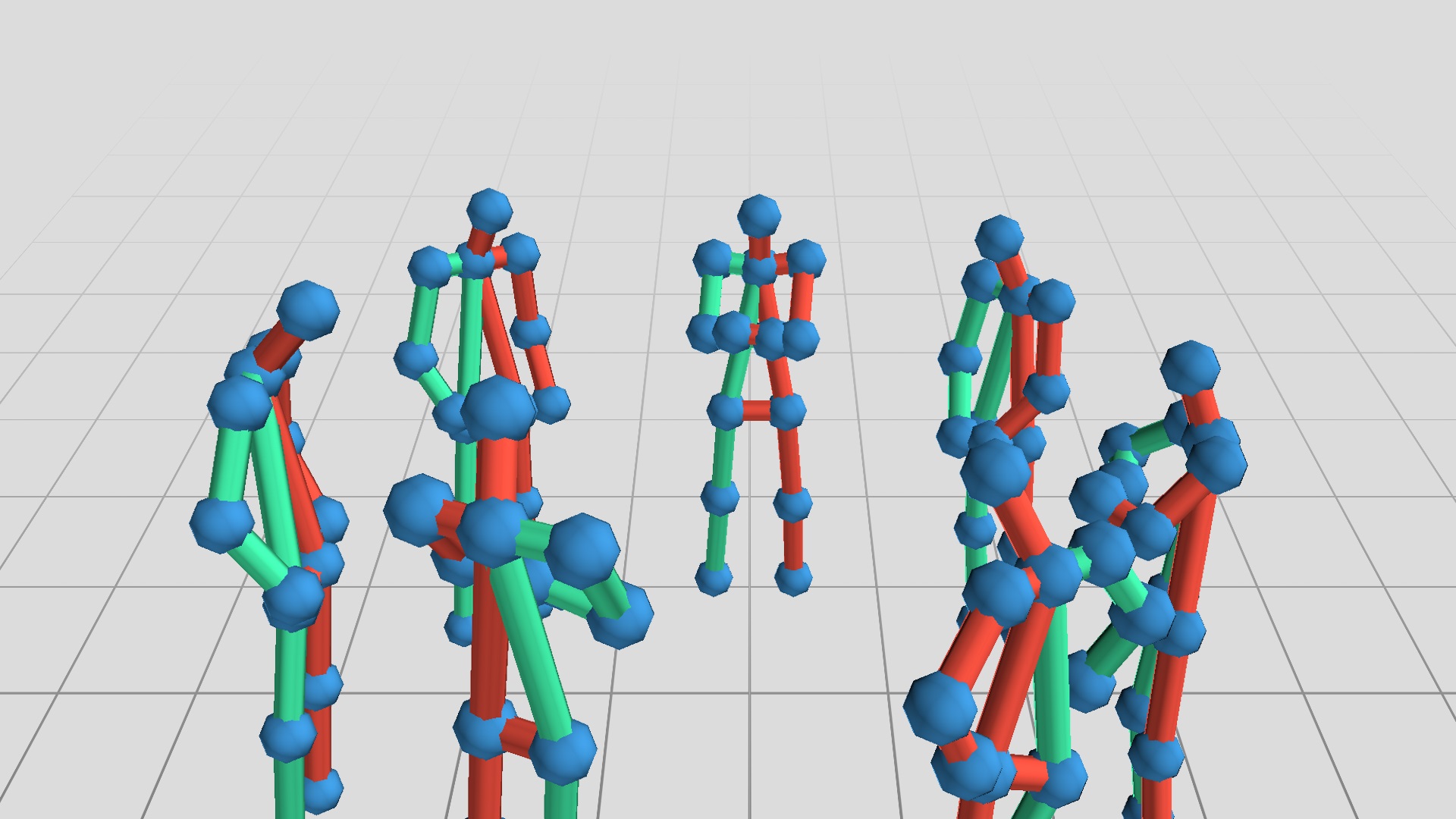}
\includegraphics[width=0.24\textwidth]{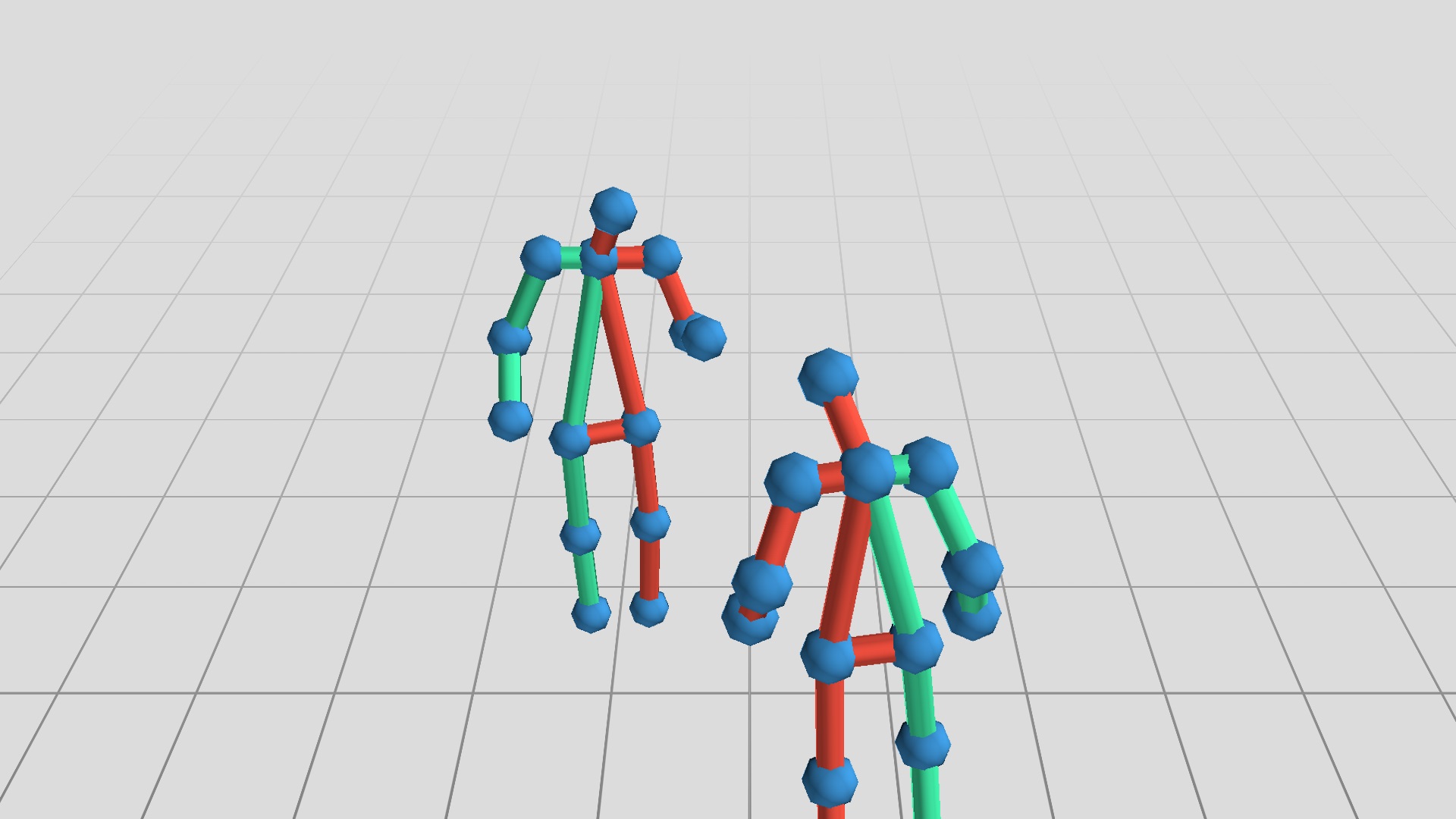}
\includegraphics[width=0.24\textwidth]{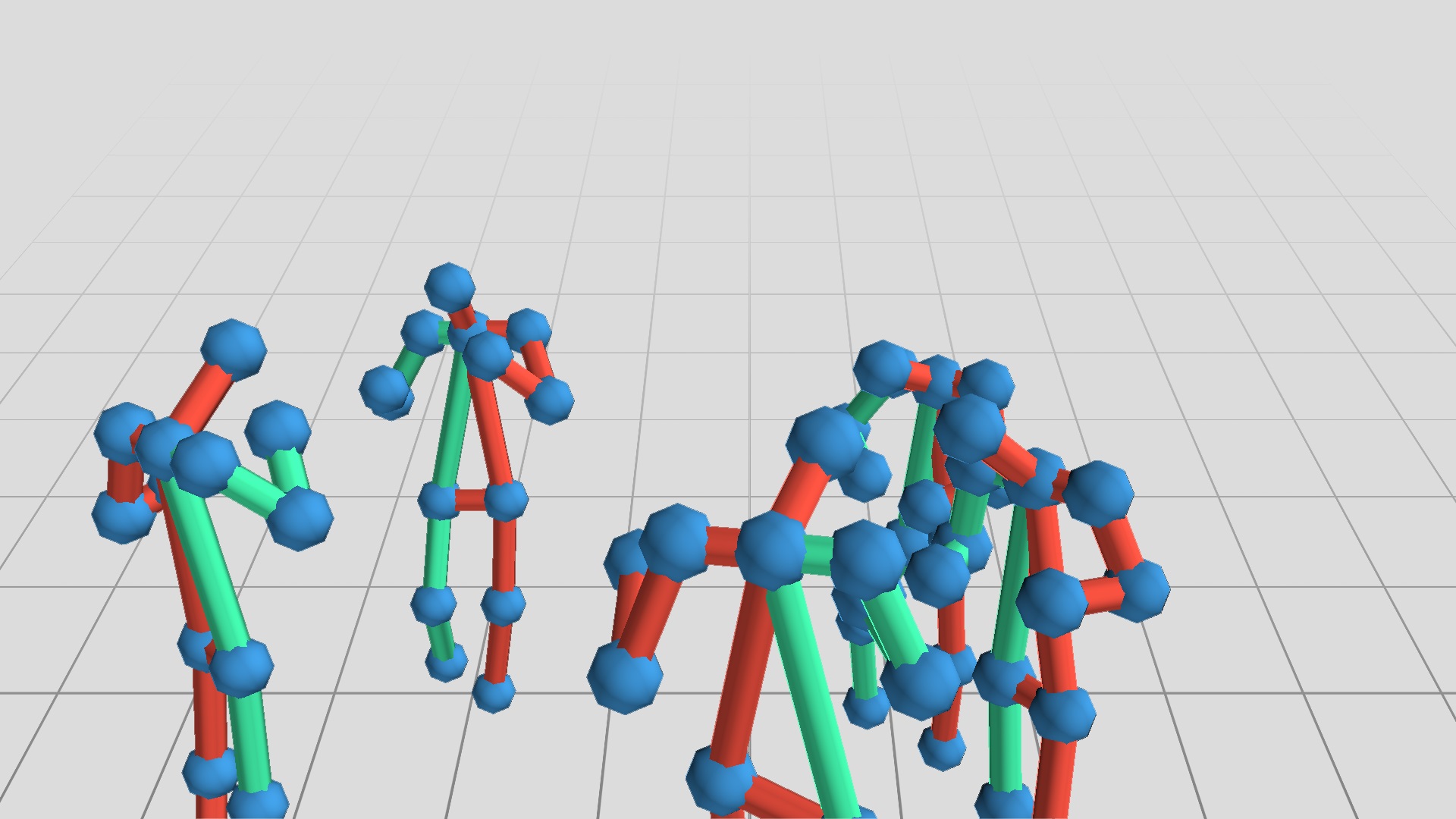}
\includegraphics[width=0.24\textwidth]{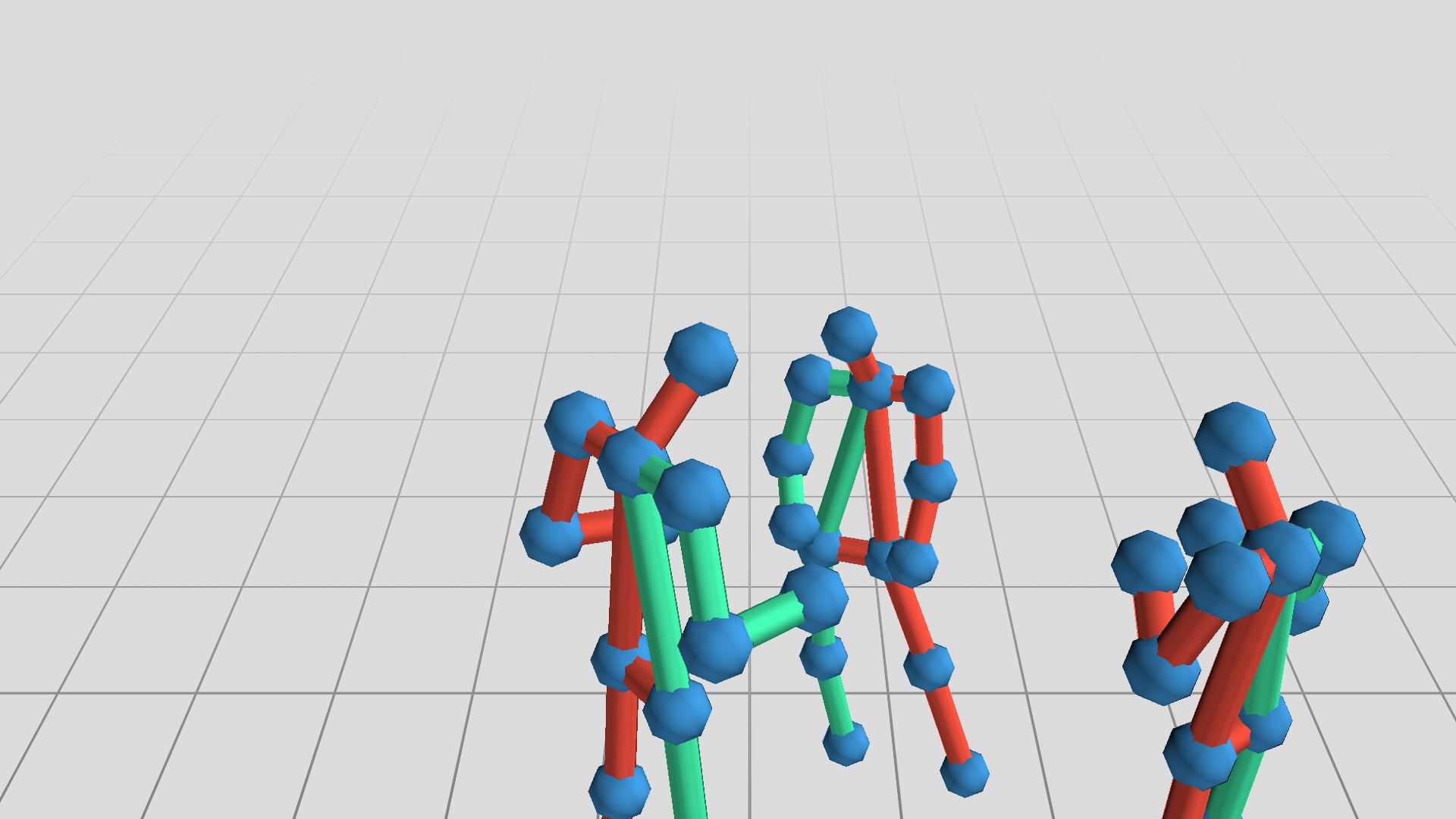}
\end{center}
\caption{Qualitative results of LoCO$^{(2)}+$ on the JTA and Panoptic datasets. We show both the 3D poses (JTA: 2\textsuperscript{nd} row, Panoptic: 4\textsuperscript{th} row) and the corresponding 2D versions re-projected on the image plane (JTA: 1\textsuperscript{st} row, Panoptic: 3\textsuperscript{rd} row)}
\label{fig:res_jta_pan}
\end{figure*}

\subsection{HPE Experiments on JTA Dataset}
\noindent
On the JTA dataset we compared LoCO against the Location Maps based approaches of \cite{mehta2018single,mehta2017vnect}. Currently the Location Maps representation is the most relevant alternative to volumetric heatmaps to approach the 3D HPE task in a bottom-up fashion and therefore represents our main competitor. 

A Location Maps is a per-joint feature channel that stores the 3D coordinate x, y, or z at the joint 2D pixel location. For each joint there are three location-maps and the 2D heatmap. The 2D heatmap encodes the pixel location of the joint as a confidence map in the image plane. The 3D position of a joint can then be obtained from its Location Map at the 2D pixel location of the joint. For a fair comparison, we utilized the same network (Inception v3 + \textit{f}-c2d) to directly predict the Location Maps. The very low F1 score demonstrate that Location Maps are not suitable for images with multiple overlapping people, not being able to effectively handle the challenging situations peculiar of crowded surveillance scenarios (see Tab.~\ref{tab:cp_exp_jta}).

Additionally, we report a comparison with a strong top-down baseline that uses YOLOv3 \cite{redmon2018yolov3} for the people detection part and \cite{martinez2017simple} as the single-person pose estimator. \cite{martinez2017simple}, like almost all single person methods, provides root-relative joint coordinates and not the absolute 3D position. We thus performed the 3D alignment according to \cite{LCRNet++} by minimizing the distance between 2D pose and re projected 3D pose. We outperform this top-down pipeline by a large margin in terms of F1-score, while being significantly faster; LoCO is able to process Full HD images with more than 50 people at 8 FPS on a Tesla V100 GPU, while the top-down baseline runs at an average of 0.5 FPS (16 times slower). The recall gap is mostly due to the fact that the detection phase in top-down approaches usually miss overlapped or partially occluded people on the crowded JTA scenes.

Finally, we compared against an end-to-end model trained to directly predict the volumetric heatmaps without compression (``Uncompr. Volumetric Heatmaps" in Tab.~\ref{tab:cp_exp_jta}). Specifically, we stacked the Code Predictor and the VHA$^{(2)}$'s decoder and trained it in an end-to-end fashion. Our technique outperforms this version at every compression rate. In fact, the sparseness of the target makes it difficult to effectively exploit the redundancy of body poses in the ground truth annotation leading to a more complex training phase.

We point out that LoCO$^{(2)}+$ obtains by far the best result in terms of F1-score compared to all evaluated approaches and baselines, thus demonstrating the effectiveness of our method. Moreover, the best result has been obtained using the VHA$^{(2)}$'s mapping, which seemingly exhibits the best compromise between information preserved and density of representation. It is also very interesting to note that the upper bound for Volumetric Heatmaps is much higher than that of Location Maps (last two rows of Tab.~\ref{tab:cp_exp_jta}), highlighting the superiority of volumetric heatmaps in crowded scenarios. It is finally worth noticing that LoCO$^{(1)}+$ and LoCO$^{(3)}+$ obtain very close results, indicating that an extremely lossy compression can lead to a poor solution as much as utilizing a too sparse and oversized representation.

Following the protocol in \cite{fabbri2018learning}, we trained all our models (and those with Location Maps) on the 256 sequences of the JTA training set and tested our complete pipeline only on every 10\textsuperscript{th} frame of the 128 test sequences.
Qualitative results are presented in Fig.~\ref{fig:res_jta_pan}.

\subsection{HPE Experiments on Panoptic Dataset}
\noindent
Here we propose a comparison between LoCO and three strong multi-person approaches \cite{zanfir2018deep, Zanfir_2018_CVPR, popa2017deep} on CMU Panoptic following the test protocol defined in \cite{Zanfir_2018_CVPR}. The results, shown in the Tab.~\ref{tab:cp_exp_panoptic}, are divided by action type and are expressed in terms of Mean Per Joint Position Error (MPJPE). MPJPE is calculated by firstly associating predicted and ground truth poses, by means of a simple Hungarian algorithm. In the Tab.~\ref{tab:cp_exp_panoptic} we also report the F1-score: the solely MPJPE metric is not meaningful as it does not take into account missing detections or false positive predictions.

The obtained results show the advantages of using volumetric heatmaps for 3D HPE, as LoCO$^{(2)}+$ achieves the best result in terms of average MPJPE on the Panoptic test set. For the sake of fairness, we also tested on the no longer maintained ``mafia'' sequence. However, the older version of the dataset utilizes a different convention for the joint positions. This, in fact, is reflected by the worst performance in that sequence only. Once again, the best trade-off is obtained using VHA$^{(2)}$, due to VHA$^{(3)}$'s mapping partial loss of information. The GT upper bound in Tab.~\ref{tab:cp_exp_panoptic} further demonstrate the potential of our representation.
% ...
Qualitative results are presented in Fig.~\ref{fig:res_jta_pan}.

\begin{table}[]
\centering
\resizebox{\linewidth}{!}{
\begin{tabular}{ccccccc}
\multicolumn{1}{l}{} & \multicolumn{5}{c}{MPJPE {[}mm{]}}      &       \\[1px] \cline{2-6}\\[-0.95em]
                     & Haggl. & Mafia & Ultim. & Pizza & Mean  &  F1   \\[1px] 
\hline \\[-0.95em]
\cite{popa2017deep}     & 218  & 187 & 194  & 221 & 203 &  -    \\
\cite{Zanfir_2018_CVPR} & 140  & 166 & 151  & 156 & 153 &  -    \\
\cite{zanfir2018deep}   & 72   & \textbf{79}  & 67   & 94  & 72  &  -    \\
LoCO$^{(2)}+$       & \textbf{45}   & 95  & \textbf{58}   & \textbf{79}  & \textbf{69}  &  89.21   \\
LoCO$^{(3)}+$       & 48   & 105  & 63   & 91  & 77  &  87.87   \\ 
\hline \\[-1.1em]
GT                      & 9    & 12   & 9    & 9   & 10   &  100   \\ 
\hline
\end{tabular}}
\caption{Comparison on the CMU Panoptic dataset. Results are shown in terms of MPJPE [mm] and F1 detection score. 
Last row: results with ground truth volumetric heatmaps}
\label{tab:cp_exp_panoptic}
\end{table}
\begin{table}[t]
\small
\centering
\begin{tabular}{clccccc}
& method & \textit{N} & P1 & P1 (a) & P2 & P2 (a) \\[0.05em]
\hline
    \\[-1em]
\multirow{4}{*}{\rotatebox[origin=c]{90}{top-down}} 
& Rogez \etal \cite{LCRNet} & 13 & 
    63.2 & 53.4 & 87.7 & 71.6
    \\    
& Dabral \etal \cite{dabral2018learning} & 16 & 
    - & - & - & 65.2
    \\
& Rogez \etal \cite{LCRNet++} & 13 & 
    54.6 & 45.8 & 65.4 & 54.3
    \\
& Moon \etal \cite{moon2019camera} & 17 & 
    \textbf{35.2} & \textbf{34.0} & \textbf{54.4} & 53.3
    \\
\hline
\multirow{4}{*}{\rotatebox[origin=c]{90}{bottom-up}} 
& Mehta \etal \cite{mehta2017vnect} & 17 & 
    - & - & 80.5 & -
    \\
& Mehta \etal \cite{mehta2018single} & 17 & 
    - & - & 69.9 & -
    \\
& LoCO$^{(2)}+$ & 14 & 
    84.0 & 75.4 & 96.6 & 77.1
    \\
& LoCO$^{(3)}+$ & 14 & 
    51.1 & 43.4 & 61.0 & \textbf{49.1}
    \\
\hline
\\[-1em]
\multicolumn{2}{c}{GT Vol. Heatmaps} 
 & 14 &
    15.6 & 14.9 & 15.0 & 14.3
\\
\hline
\end{tabular}%
\vspace{2mm}
\caption{Comparison on the Human3.6m dataset in terms of average MPJPE [mm]. ``(a)" indicates the addition of rigid alignment to the test protocol; $N$ is the number of joints considered by the method. Last row: results with ground truth volumetric heatmaps}
\label{tab:cp_human_exp}
\end{table}

\subsection{HPE Experiments on Human3.6m Dataset}
\noindent
In analogy with previous experiments, we tested LoCO on Human3.6m. Unlike most existing approaches, we apply our multi-person method as it is, without exploiting the knowledge of the single-person nature of the dataset, as we want to demonstrate its effectiveness even in this simpler context. Results, with and without rigid alignement, are reported in terms of MPJPE following the P1 and P2 protocols. In the P1 protocol, six subjects (S1, S5, S6, S7, S8 and S9) are used for training and every 64\textsuperscript{th} frame of subject S11/camera 2 is used for testing. For the P2 protocol, all the frames from subjects S9 and S11 are used for testing and only S1, S5, S6, S7 and S8 are used for training. 

\begin{figure}
\begin{center}
\includegraphics[trim={5em 4em 5em 3em},clip,width=0.24\linewidth]{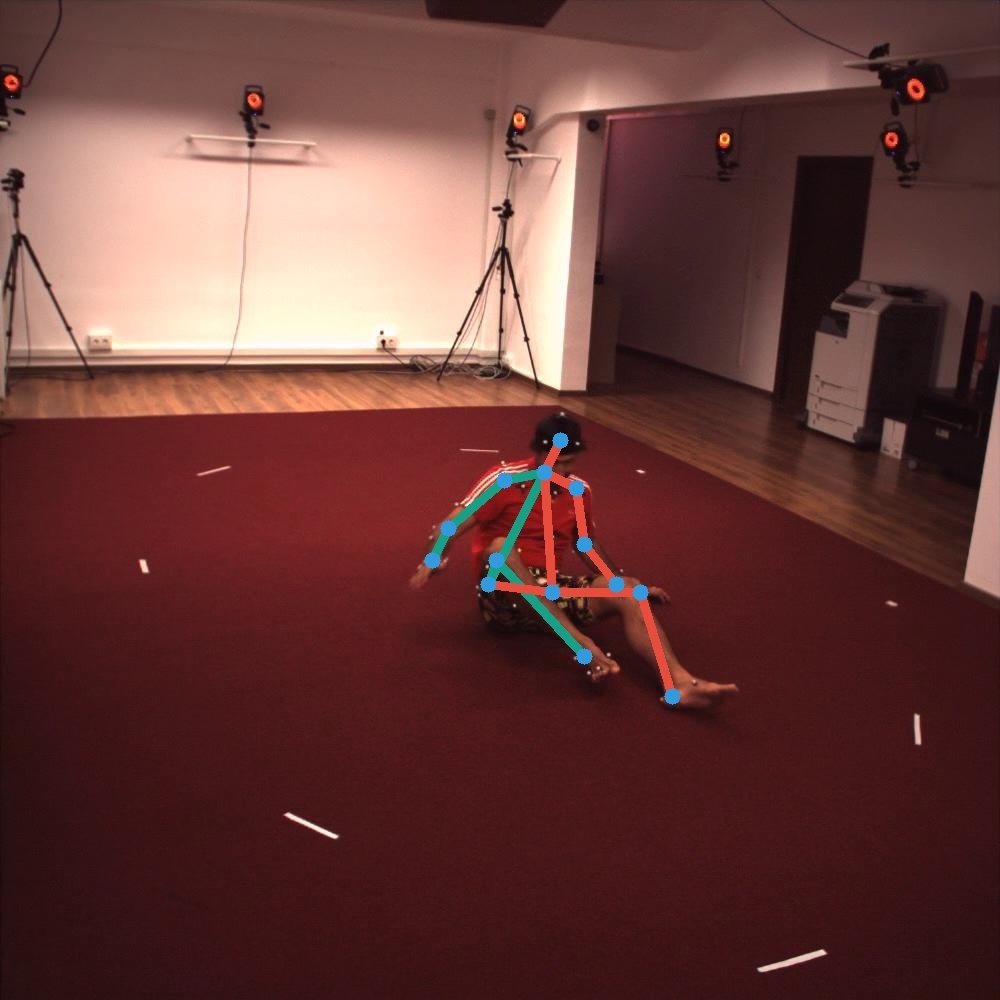}
\includegraphics[trim={5em 4em 5em 3em},clip,width=0.24\linewidth]{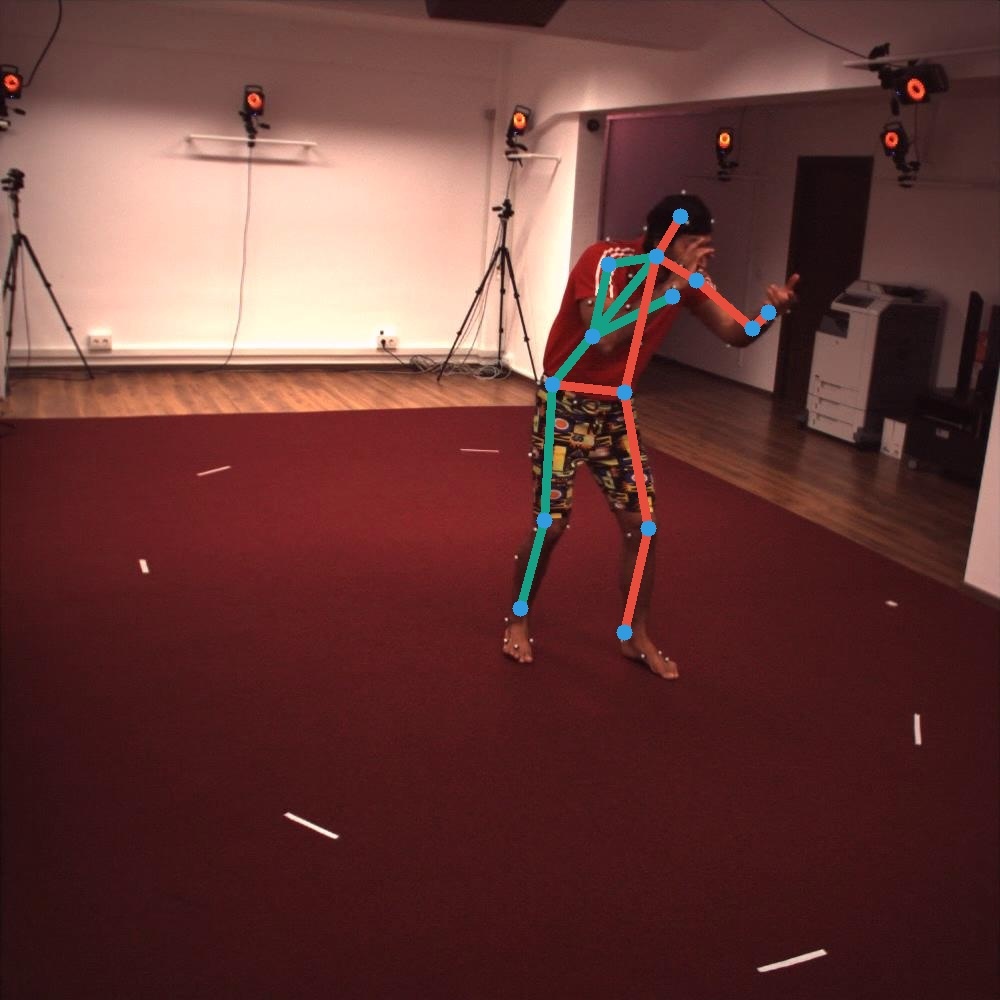}
\includegraphics[trim={5em 4em 5em 3em},clip,width=0.24\linewidth]{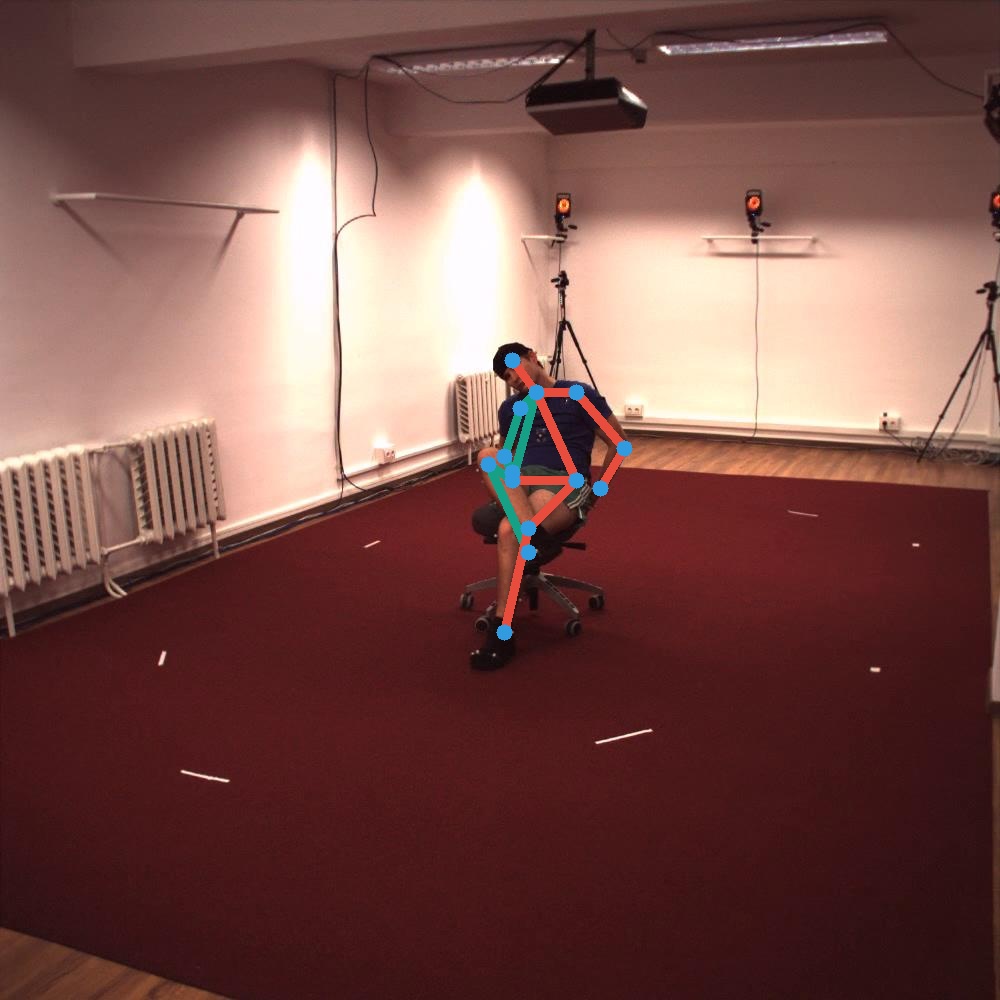}
\includegraphics[trim={5em 4em 5em 3em},clip,width=0.24\linewidth]{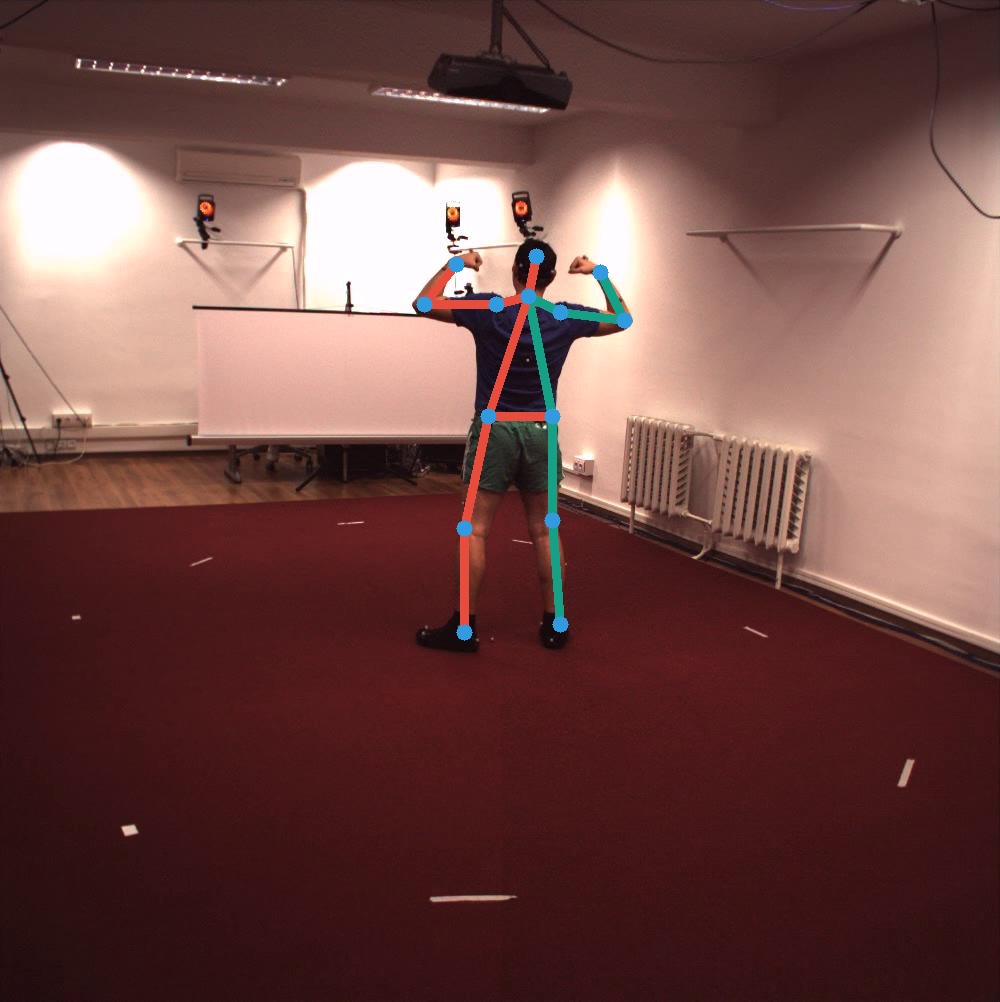}
\\
\includegraphics[trim={5em 4em 5em 3em},clip,width=0.24\linewidth]{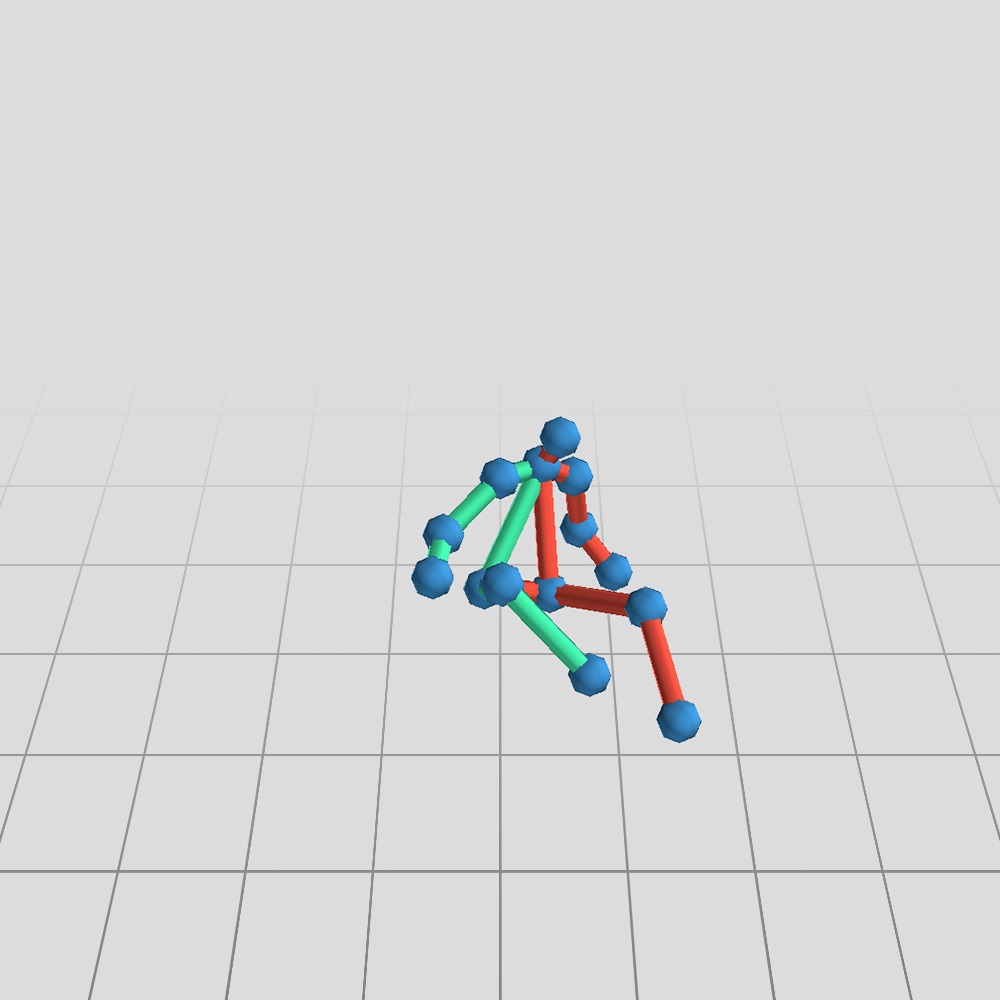}
\includegraphics[trim={5em 4em 5em 3em},clip,width=0.24\linewidth]{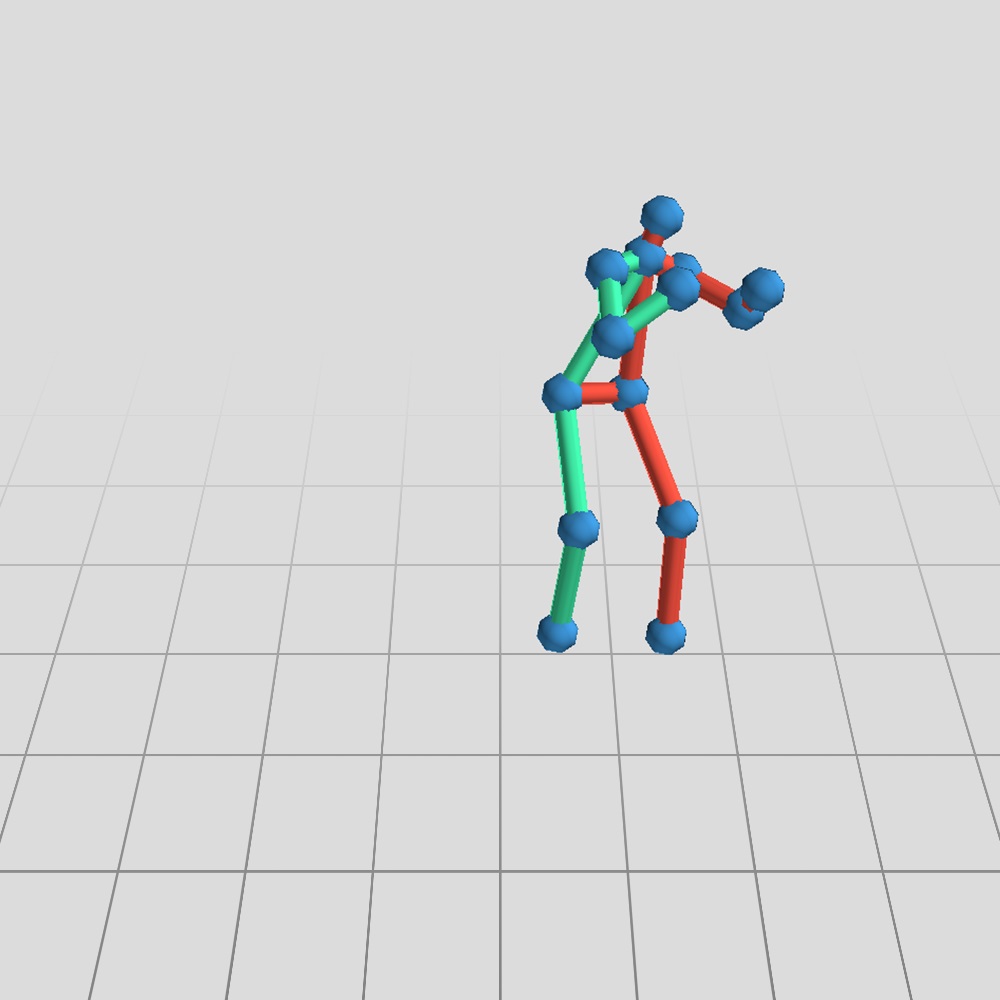}
\includegraphics[trim={5em 4em 5em 3em},clip,width=0.24\linewidth]{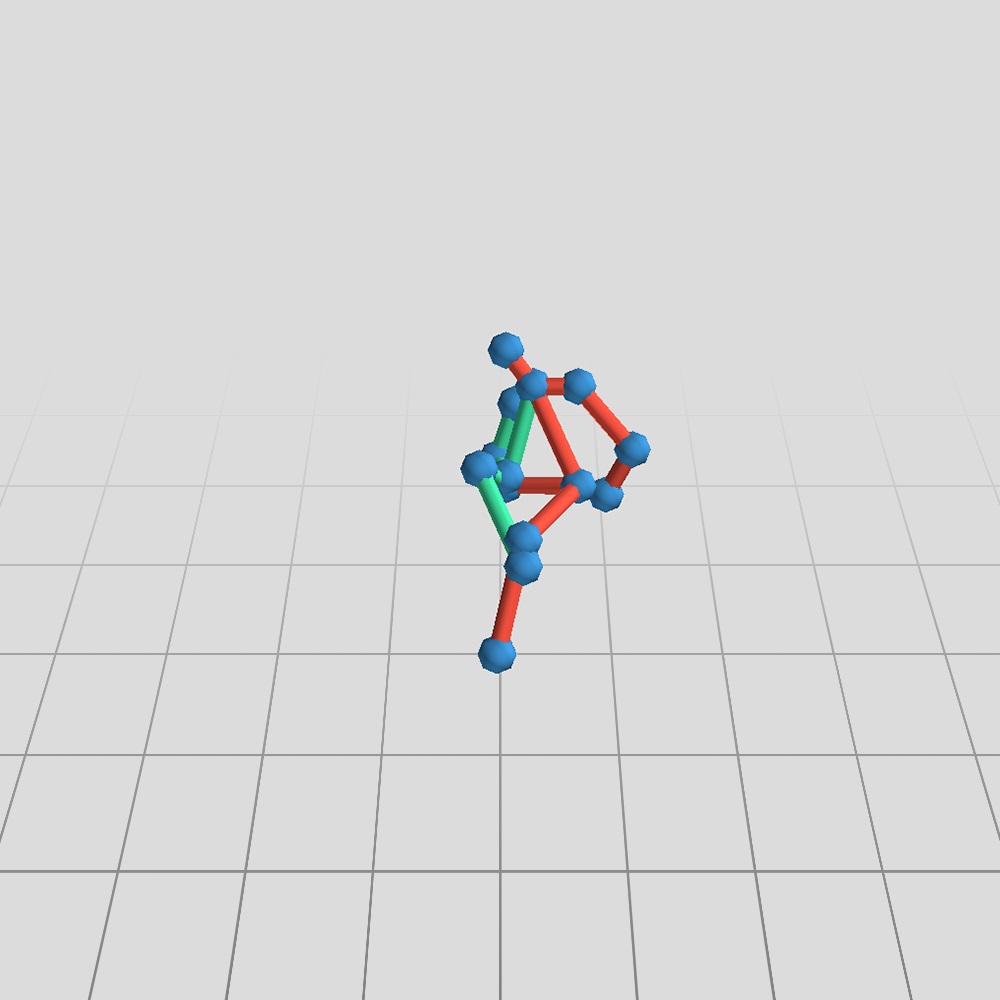}
\includegraphics[trim={5em 4em 5em 3em},clip,width=0.24\linewidth]{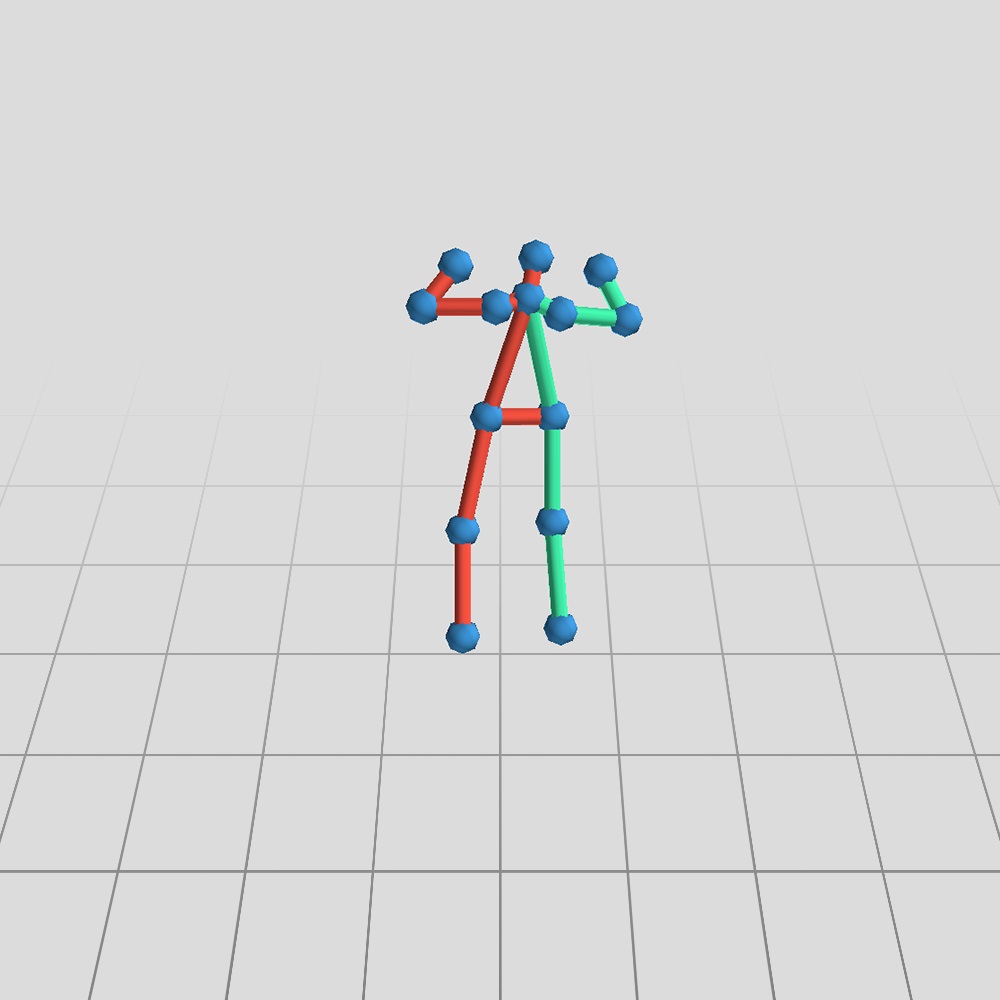}
\end{center}
    \caption{Qualitative results of LoCO$^{(3)}+$ on the Human3.6m dataset}
\label{fig:res_human}
\end{figure}

Tab.~\ref{tab:cp_human_exp} shows a comparison with recent state-of-the-art multi-person methods, showing that our method is well suited even in the single person context, as LoCO$^{(3)}+$ achieves state of the art results among bottom up methods. Note that, although Moon \etal reports better numerical performance, they leverage additional data for training and evaluate on a more redundant set of joints containing pelvis, torso and neck. It is worth noticing that LoCO$^{(3)}+$ performs substantially better than LoCO$^{(2)}+$, demonstrating that a smaller representation is preferred when the same amount of information is preserved (99.7 and 100.0 F1@0vx respectively on VHA$^{(3)}$ and VHA$^{(2)}$). Qualitative results are presented in Fig.~\ref{fig:res_human}.

\section{Discussion and Conclusions}
\noindent
In conclusion, we presented a single-shot bottom-up approach for multi-person 3D HPE suitable for both crowded surveillance scenarios and for simpler, even single person, contexts without any changes. Our LoCO approach allows us to exploit volumetric heatmaps as a ground truth representation for the 3D HPE task. Instead, without compression, this would lead to a sparse and extremely high dimensional output space with consequences on both the network size and the stability of the training procedure. In comparison with top-down approaches, we removed the dependency on the people detector stage, hence gaining both in terms of robustness and assuring a constant processing time at the increasing of people in the scene. The experiments show state-of-the-art performance on all the considered datasets. We also believe that this new simple compression strategy can foster future research by enabling the full potential of the volumetric heatmap representation in contexts where it was previously intractable. 

\section*{Acknowledgments}
\noindent
This work was supported by Panasonic Corporation and by the Italian Ministry of Education, Universities and Research under the project COSMOS PRIN 2015 programme 201548C5NT.

{\small
\bibliographystyle{ieee_fullname}
\bibliography{egbib}
}

\clearpage



\section*{Supplementary material}

# Supplementary Material

Matteo Fabbri[1*]  Fabio Lanzi[1]  Simone Calderara[1]  Stefano Alletto[2]  Rita Cucchiara[1]

[1]University of Modena and Reggio Emilia  [2]Panasonic R&D Company of America

{name.surname}@unimore.it  {name.surname}@us.panasonic.com

## 1. Implementation Details

For the sake of reproducibility, in this section we illustrate the architectures of the Code Predictor and the Pose Refiner modules of our LoCO pipeline.

**Code Predictor**  Inspired by [4], our method simply adds a few convolutional layers ($f$-c2d) to the last convolution stage of a backbone network. Tab. 1 reports the detailed structures of the various $f$-c2d blocks utilized in our experiments. *ConvTr2D* refers to transposed 2D convolutions while *Conv2D* refers to simple 2D convolutions. For each layer, we provide: number of input channels, number of output channels, kernel size and stride. In all the proposed experiments we utilized InceptionV3 [3] pretrained on Imagenet [1] as backbone architecture.

**Pose Refiner**  The structure of the Pose Refiner is shown in Fig. 1. It is a simple network composed by three fully connected layers with ReLU activation followed by a skip connection. Input and output are normalized root-relative representations of a single 3D pose, with values in range $[0, 1]$. During training, Gaussian noise (mean: $0m$, variance: $0.08m$) is applied to the input pose while some joints are randomly removed with probability 0.1. The removed joints are coded with a default value of $(-1, -1, -1)$.

## 2. Volumetric Heatmap Spaces

In our experiments, we defined our Volumetric Heatmap representation according to two different pseudo-3D spaces, depending on which dataset we used:

- The first space is defined as $S_1 = D \times H' \times W'$, where $H'$ and $W'$ are the height and width, downsampled by a actor of 8, of the image plane and $D$ is the maximum distance from the camera in meters, quantized with 316 bins. We adopted $S_1$ for JTA.

|  | layer | in ch. | out ch. | ker. | str. |
|---|---|---|---|---|---|
| LoCO$^{(1)}$ | ConvTr2D→ReLU | $F$ | 1024 | 4 | 2 |
|  | ConvTr2D→ReLU | 1024 | 512 | 4 | 2 |
|  | Conv2D→ReLU | 512 | 256 | 4 | 1 |
|  | Conv2D | 256 | 158 | 1 | 1 |
| LoCO$^{(2)}$ | ConvTr2D→ReLU | $F$ | 1024 | 4 | 2 |
|  | Conv2D→ReLU | 1024 | 512 | 4 | 1 |
|  | Conv2D→ReLU | 512 | 256 | 4 | 1 |
|  | Conv2D | 256 | 79 | 1 | 1 |
| LoCO$^{(3)}$ | Conv2D→ReLU | $F$ | 1024 | 4 | 1 |
|  | Conv2D→ReLU | 1024 | 512 | 4 | 1 |
|  | Conv2D→ReLU | 512 | 256 | 4 | 1 |
|  | Conv2D | 256 | 39 | 1 | 1 |

Table 1. Structure of the three $f$-c2d block variants of the Code Predictor used for our HPE experiments. $F$ represents the number of output channels of the exploited feature extractor. In all our experiments we used Inception v3 with $F = 2048$

- The second space is defined as $S_2 = Z \times H' \times W'$, where $H'$ and $W'$ are defined as in $S_1$, and $Z$ is the maximum **z** axis value of the real 3D space in the standard coordinate system centered to the camera. $Z$ is expressed in meters and quantized with 316 bins. We adopted $S_2$ for Panoptic and Human3.6m.

Although the difference between these two spaces is minimal, we adopted $S_1$ for JTA because this dataset already provide a maximum camera distance, which is 100 meters.

## 3. Detection Experiments

To show how our LoCO approach can be effectively adopted also for the detection task in crowded scenarios under heavy occlusions, we have tested our system in terms of 2D people detection comparing it with YOLOv3 [2] on the JTA test set. Using LoCO, we predict 3D poses and project them on 2D bounding boxes using the camera intrinsic parameters.

In terms of precision, recall, and F1 (with the bounding

---

* Work done while interning at Panasonic R&D Company of America

1

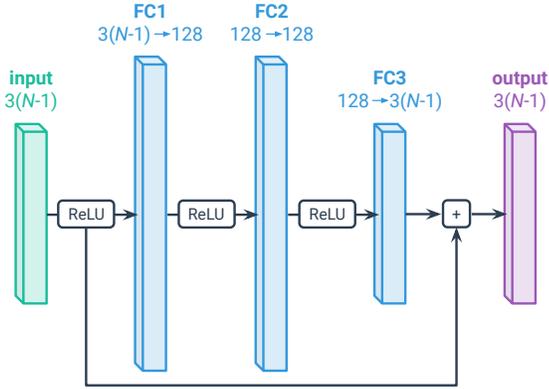

Figure 1. Structure of our Pose Refiner: 3 fully connected layers with ReLU activation followed by a skip connection. $N$ is the number of joints. In all our experiments we considered $N = 14$.

box IoU threshold at 0.5), using our LoCO$^{(2)}$+ trained on JTA, we get 81.94, 69.73, and 75.39 respectively; with out of the box YOLOv3, instead, we obtain 99.12, 30.81 and 44.50.

Although our model is less precise than YOLOv3 (around -20%), it surpasses it by a large margin (around +40%) in terms of recall, resulting in an F1-score that is clearly in our favor (almost +30%). The scenes in JTA, in fact, are extremely crowded and present a very high percentage of occlusion with multiple overlapping people. It is not easy for a detector to handle situations of this type, while a part-based bottom-up method is much less affected by this problem.

## 4. Skeleton Grouping Details

Let's consider $K$ different type of joints and $N_0, \ldots, N_{K-1}$ number of detections for each joint type. Given $N_0$ predicted heads, $\mathbf{j}_{0,0}, \ldots, \mathbf{j}_{0,N_0-1} \in \mathbb{R}^3$, and $N_k, k \in [1, K-1]$ predicted joints of another type, $\mathbf{j}_{k,0}, \ldots, \mathbf{j}_{k,N_k-1} \in \mathbb{R}^3$, we define $K-1$ cost matrices, $\mathfrak{D}_1, \ldots, \mathfrak{D}_{k-1}$, as follows: $\mathfrak{D}_k : \{0, \ldots, N_0-1\} \times \{0, \ldots, N_k-1\} \to \mathbb{R}$ where each element $d_{a,b}$ is defined as

$$d_{a,b} = \begin{cases} \|j_{0,a} - j_{k,b}\| & \text{if } \|j_{0,a} - j_{k,b}\| \leq 1.5 \cdot \tau_k \\ +\infty & \text{otherwise} \end{cases} \quad (1)$$

The threshold $\tau_k$ in (1) is the maximum distance between a head and a joint of type $k$ (belonging to the same person) on the entire training set. For each $k = 1, \ldots, K-1$, joint-head associations are made with the Hungarian algorithm using $\mathfrak{D}_k$ as cost matrix. The same joint-grouping procedure is applied on both multi-person datasets. By removing the anatomical constraints, results on Panoptic show an MPJPE degradation of about 9 millimeters while on JTA, no degradation in terms of metric has been observed.

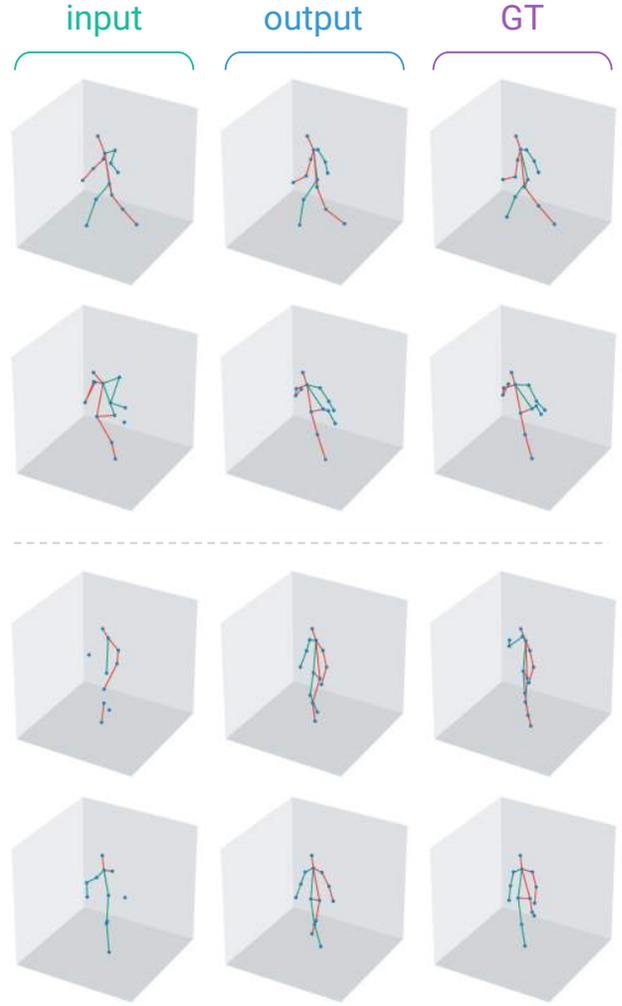

Figure 2. Qualitative results of our Pose Refiner model on the JTA dataset. $1^{st}$ and $2^{nd}$ rows: examples where the output is anatomically plausible and consistent with the ground truth; $3^{rd}$ and $4^{th}$ rows: examples where the output is anatomically plausible, but inconsistent with the ground truth.

## 5. Additional Qualitative Results

We report the results of our Pose Refiner on JTA. When the input pose fed to our Pose Refiner has few missing joints and position errors, the reconstruction is consistent with the ground truth pose (Fig. 2 - first two rows). Conversely, when the input pose is more degraded, the reconstruction is still plausible, but not always coherent with the ground truth (Fig. 2 - last two rows). Additionally, Fig. 3 depicts some qualitative results of LoCO on JTA, CMU Panoptic and Human3.6m. Our method can be applied to both crowded scenarios and single person contexts, displaying good generalization capabilities in a wide range of contexts.

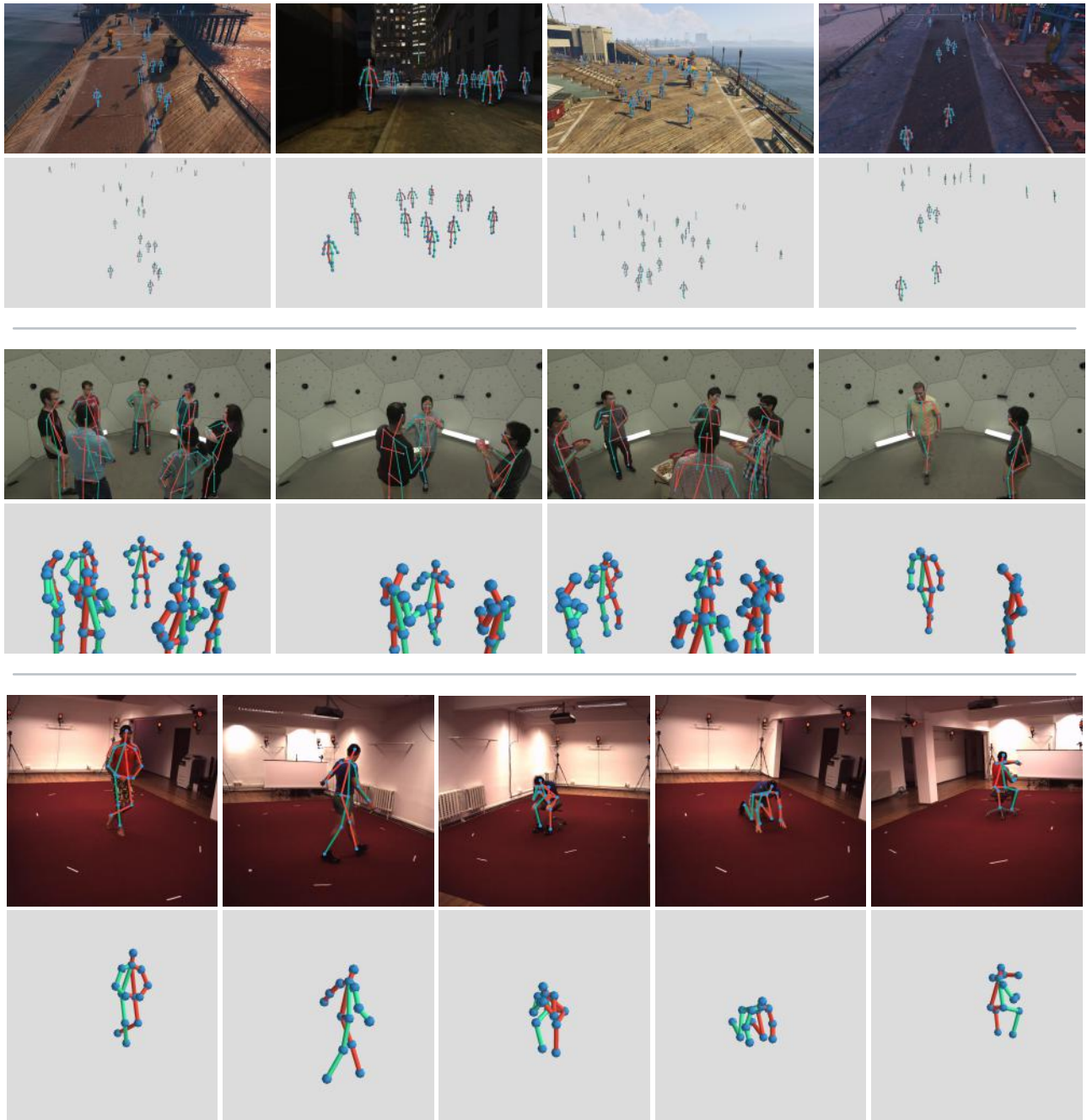

Figure 3. Additional qualitative results of our LoCO approach. 1$^{st}$ and 2$^{nd}$ rows: result of LoCO$^{(2)}$+ on the JTA dataset; 3$^{rd}$ and 4$^{th}$ rows: result of LoCO$^{(2)}$+ on the CMU Panoptic dataset; 5$^{th}$ and 6$^{th}$ rows: result of LoCO$^{(3)}$+ on the Human3.6m dataset



\end{document}